\documentclass[journal]{IEEEtran}
\usepackage{amsmath,graphicx}
\usepackage{multirow}
\usepackage{hhline}
\usepackage{multicol, blindtext}
\usepackage{array}
\usepackage{placeins}
\usepackage{float}
\usepackage{color}
\usepackage{makecell}
\usepackage{array}
\usepackage{lipsum}

\usepackage{cite}

\begin{document}

\title{Unsupervised Iterative Deep Learning of Speech Features and Acoustic Tokens with Applications to Spoken Term Detection}


\author{Cheng-Tao~Chung, Cheng-Yu Tsai, Chia-Hsiang Liu and
        Lin-Shan~Lee,~\IEEEmembership{Fellow,~IEEE}}
\maketitle

\thispagestyle{plain}
\pagestyle{plain}

\begin{abstract}
In this paper we aim to automatically discover high quality frame-level speech features and acoustic tokens directly from unlabeled speech data. A Multi-granular Acoustic Tokenizer (MAT) was proposed for automatic discovery of multiple sets of acoustic tokens from the given corpus. Each acoustic token set is specified by a set of hyperparameters describing the model configuration. These different sets of acoustic tokens carry different characteristics for the given corpus and the language behind, thus can be mutually reinforced. The multiple sets of token labels are then used as the targets of a Multi-target Deep Neural Network (MDNN) trained on frame-level acoustic features. Bottleneck features extracted from the MDNN are then used as the feedback input to the MAT and the MDNN itself in the next iteration. The multi-granular acoustic token sets and the frame-level speech features can be iteratively optimized in the iterative deep learning framework. We call this framework the Multi-granular Acoustic Tokenizing Deep Neural Network (MAT-DNN). The results were evaluated using the metrics and corpora defined in the Zero Resource Speech Challenge organized at Interspeech 2015, and improved performance was obtained with a set of experiments of query-by-example spoken term detection on the same corpora. Visualization for the discovered tokens against the English phonemes was also shown.
\end{abstract}


\section{Introduction}
%
%
\IEEEPARstart{I}{n} the era of big data, huge quantities of raw speech data is easy to obtain, but annotated speech data remain hard to acquire.  
%
This leads to the increased importance of unsupervised learning scenarios where annotated data is not required, such as Query-by-Example Spoken Term Detection (QbE-STD). 
With the dominant paradigm of automatic speech recognition (ASR) technologies being supervised learning \cite{hinton2012deep}, such a scenario is still a relatively less explored area.

This paper focuses on the unsupervised learning scenario without any annotated data, with a goal to produce alternative representations of the raw speech signal that better describe the characteristics of the underlying unknown linguistics. 
Today the two most popular forms of speech signal representation are either a sequence of real-valued frame-level feature vectors (like Mel-Frequency Cepstral Coefficients (MFCC) or spectrogram), or a sequence of discrete tokens (like words or phonemes). 
The former may be obtained in an unsupervised way, but usually only carrying relatively primitive information; the latter may be more related to the semantics, but usually only obtainable with supervised approaches requiring annotated data.
%
{\color{black}
Likewise, the initial effort of zero resource speech processing technologies primarily focused on extracting either frame-level features\cite{szoke2015copingwith,leung2016toward,chen2016unsupervised,yang2014intrinsic,wang2015acoustic,renshaw2015comparison,zhang2013unsupervised,huang2015rapid} or discrete tokens \cite{jansen2012indexing,jansen2011towards,gish2009unsupervised,chung2013unsupervised,chung2014unsupervised,chung2015enhancing,li2013towards} out of a given corpus.
}

For the task of unsupervised discovery of discrete acoustic tokens, most effort discovered only one level of phone-like acoustic tokens.
However, it is well known that multiple levels of tokens exist in speech with varying lengths, such as phones, syllables, words and phrases.
We call this the Temporal Granularity of the token sets in this work. 
On the other hand, it is also well known that multiple levels of acoustic token representations exist in speech such as speaker-independent phonemes, gender-dependent phonemes, and speaker dependent phonemes are examples of different ways of clustering the acoustic units. 
We call this the Phonetic Granularity of the token sets in this work.
In our previous work \cite{chung2014unsupervised,chung2015enhancing}, we proposed the concept of a granularity space, in which the temporal and phonetic granularities mentioned above are simply two dimensions, and every token representation for a given corpus is simply mapped onto a point in this space. 
Different token representations based on token sets mapped to different points on the granularity space were shown to carry complementary information describing the characteristics of the language behind when jointly considered.
This was the first time the concept of multi-dimensional granularity space was introduced for unsupervised token discovery.
These multiple sets of acoustic tokens based on the granularity space \cite{chung2014unsupervised,chung2015enhancing} were trained using approaches under the unsupervised Gaussian Mixture Model-Hidden Markov Model (GMM-HMM) framework. 
In recent years, deep learning approaches have been applied in almost all speech processing tasks. 
It is therefore natural to believe that the unsupervised approaches discussed here can also be benefited by deep learning approaches.
%
However, the Deep Neural Network (DNN) is a discriminative model and is more difficult to be applied to the unsupervised scenario here. 
This implies we need a completely novel deep learning framework for such unsupervised scenario, and this is exactly the topic of this paper.

Recently, we proposed a new concept along the direction mentioned above to discover multiple acoustic token sets over the granularity space with a deep learning framework \cite{chung2015iterative}, and iteratively learn the frame-level signal representations and the acoustic tokens from a corpus iteratively.
%
%
%
%
To our knowledge this was the first time these two representations were jointly learned and optimized in a single framework under the zero resource scenario.
This framework was presented in the paper as a recursive iterative learning between a Multi-granular Acoustic Tokenizer (MAT)\footnote{In \cite{chung2015iterative}, the Multi-granular Acoustic Tokenizer was called the Multi-layered Acoustic Tokenizer.}
 and a Multi-target Deep Neural Network (MDNN).
The Multi-granular Acoustic Tokenizer (MAT) was used to generate multiple sets of acoustic tokens, each specified by a point on the granularity space representing a specific HMM configuration referred to as a level. 
The different levels of the tokens carry complementary knowledge about the corpus and the language behind \cite{chung2014unsupervised}, thus can be further mutually reinforced \cite{chung2015enhancing}. 
The multi-granular token labels generated by the MAT were then used as the training targets of a Multi-target Deep Neural Network \cite{vu2014investigating} (MDNN) to learn the frame-level bottleneck features \cite{vesely2012language} (BNFs). 
These BNFs were then used as the feedback input to both the MAT and the MDNN in the next iteration. 
The whole framework was referred to as a Multi-granular Acoustic Tokenizing Deep Neural Network (MAT-DNN). 
%

A Zero Resource Speech Challenge was organized in Interspeech 2015 \cite{versteegh2015zero} considering the extreme situation where a whole language has to be learned from scratch from a given corpus \cite{lee2012nonparametric,siu2014unsupervised,kamperunsupervised,levin2015segmental}, which was very close the scenario described here. 
In the Challenge, a whole set of evaluation metrics and training corpora were defined and organized into two tracks. 
%
Track 1 was about evaluating the frame-level speech features, while
Track 2 was about the discovering of word-like tokens from the corpus.
In both tracks the defined evaluation metrics are more general considering all different aspects without specifying any applications.
In the paper, we used the evaluation metrics and corpora defined by the challenge in our experiments for easier comparison of results.
In addition, we chose query-by-example spoken term detection to demonstrate that the discovered frame-level features and acoustic tokens work well in a real applications.  
%
%

In this work, we further extend \cite{chung2015iterative} by providing more experimental data, adding new comparisons with other baselines, and performing two additional visualization experiments.
In the explanation of the proposed approach, we added a detailed description of the initialization of the token sets.
For the experiments on the metrics defined by the Challenge, we added new results and compared our results with other participants of the Challenge that were not available at the time of submission of \cite{chung2015iterative}.
For the spoken term detection experiments, we added the comparison of our unsupervised systems with supervised phoneme recognizers on 4 different languages, and offered visualization of the performance of tokens for different granularities.
We also performed visualization experiments on tokens with different granularities by mapping the discovered tokens to words or phonemes on the TIMIT training set.

The rest of the work is organized in the following order.
%
%
The proposed approach is presented in Section \ref{sec:II}, with experimental setup in Section \ref{sec:III}.
%
%
The experimental results are described in Section \ref{sec:V}, with concluding remarks in Section \ref{sec:VI}.

\begin{figure*}[tbh]
\centerline{\includegraphics[width=1.0\textwidth]{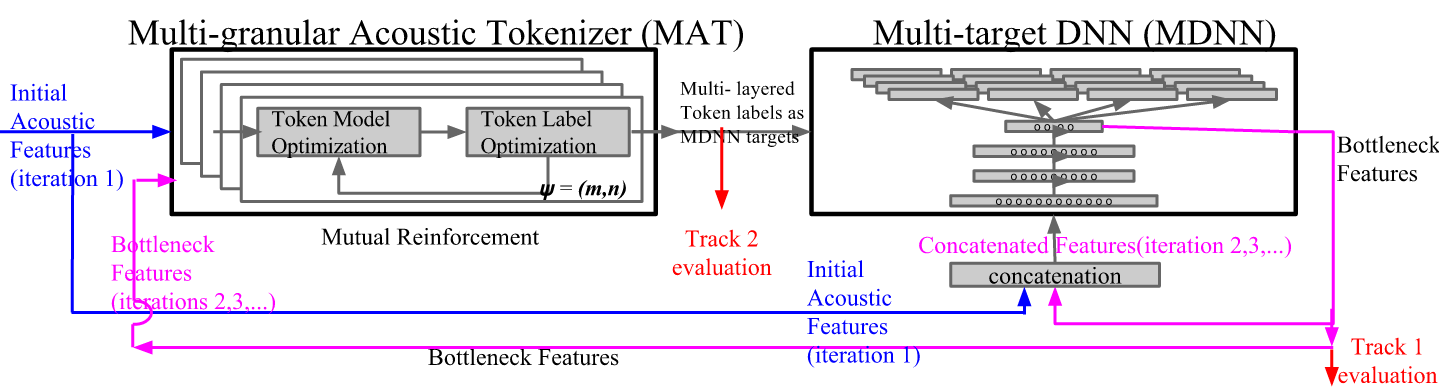}}
\caption{The proposed framework of Multi-granular Acoustic Tokenizing Deep Neural Network (MAT-DNN)}\label{fig:1}
\end{figure*}

\section{Proposed Approach}\label{sec:II}

\subsection{Overview of the proposed framework}
The overall framework of the proposed approach is shown in Fig\ref{fig:1}.
In the left part, the Multi-granular Acoustic Tokenizer (MAT) produces many sets of acoustic tokens, each describing some aspects of the given corpus.
Every token in a set is modeled by an HMM trained in an unsupervised way.
{\color{black}
Each set of the tokens is referred to as a level specified by a set of two hyperparameters $\psi=(m,n)$ describing the HMM configurations: the number of states $m$ in each acoustic token HMM, and the total number of distinct acoustic tokens $n$ during initialization. 
We will show in Section \ref{sec:2-2-2} that these hyperparameters correspond to strong physical meanings.
}
Each set of acoustic tokens for each configuration is obtained by iteratively optimizing the token models (HMMs) and the token labels on the given audio corpus, as shown in the left part of Fig. \ref {fig:1}. 
Multiple sets of the hyperparameters $\psi=(m,n)$ were selected to produce multi-granular token labels for the given audio corpus to be used as the training targets of the Multi-target Deep Neural Network (MDNN) on the right part of Fig. \ref{fig:1}, in which the knowledge carried by different token sets on different levels are fused.
Bottleneck features are then extracted from this MDNN. 
In the first iteration, some initial frame-level acoustic features (such as MFCCs) are used for both the MAT and the MDNN. 
This gives the first set of bottleneck features. 
These bottleneck features are then used as the feedback to both the MAT (to replace the initial acoustic features) and the MDNN (to be concatenated with the initial acoustic features to produce tandem features) in the second iteration. 
Such feedback can be continued iteratively. The complete framework is referred to  as Multi-granular Acoustic Tokenizing Deep Neural Network (MAT-DNN). 
The output of the MDNN (bottleneck features) is the discovered frame-level features and evaluated with the metrics defined in Track 1 of the Challenge, while the acoustic token labels at the output of the MAT are evaluated with the metrics defined in Track 2 of the Challenge. 
Both the acoustic tokens and frame-level features are then further evaluated in a real application of query-by-example spoken term detection task. 


\subsection{Multi-granular Acoustic Tokenizer(MAT)}
\subsubsection{Unsupervised Token Discovery for Each level of MAT}
\label{sec:2-2-1}
The Multi-granular Acoustic Tokenizer (MAT) in the left part of Fig. \ref{fig:1} is presented in this subsection.
In the set of hyperparameters $\psi=(m,n)$ used to define the HMM configuration for a level of token models, $m$ is the number of states per model and $n$ is the number of distinct models.
The goal here is to obtain  multiple sets of acoustic tokens in a completely unsupervised way from a given audio corpus, each defined by a set of hyperparameters $\psi=(m,n)$.
{\color{black}
It is straightforward to discover acoustic tokens from the corpus for a chosen hyperparameter set $\psi=(m,n)$ that determines the HMM configuration \cite{lee1988segment,wang2011unsupervised,jansen2011towards,gish2009unsupervised,chung2013unsupervised,siu2010improved,creutz2007unsupervised}.
}
On each level or for each set of $\psi=(m,n)$, this can be achieved by first finding an initial label set $W_0$ based on a set of assumed tokens for all features in the corpus $X$ as in (\ref{eq:1}) \cite{chung2013unsupervised}.
Then in each iteration $t$ the HMM parameters $\theta^\psi_{t}$ can be trained with the label set $W_{t-1}$ obtained in the previous iteration with EM algorithm based on maximum likelihood criterion as in (\ref{eq:2}), which is exactly the same as the supervised training of HMMs except based on the label set $W_{t-1}$ rather than the real annotations.
The new label set $W_{t}$ can then be obtained by token decoding with the obtained parameters $\theta^\psi_{t}$ using Viterbi algorithm as in (\ref{eq:3}). 
This process is also described in \cite{gish2009unsupervised,wang2015acoustic}. 
\begin{eqnarray}
W_{0}&=& \mbox{initialization} (X),\phantom{\arg \max_{\substack{\theta^\psi}}}                                           \label{eq:1} \\ 
\theta^\psi_{t} &=& \arg \max_{\substack{\theta^\psi}} P(X|\theta^\psi,W_{t-1}),             \label{eq:2} \\
W_{t} &=& \arg \max_{\substack{\omega}} P(X|\theta^\psi_{t} ,W).                        \label{eq:3}
\end{eqnarray}
The superscript $\psi$ on the model parameter set $\theta^\psi$ indicated there is a distinct set of parameters for each $\psi=(m,n)$.
The training process can be repeated with enough number of iterations until a converged set of token HMMs is obtained. 
The processes (\ref{eq:2}), (\ref{eq:3}) are respectively referred to as token model optimization and token label optimization in the left part of Fig. \ref{fig:1}.

\subsubsection{Initialization}
\begin{figure}[tb]
  \centering
    \includegraphics[width=0.9\columnwidth]{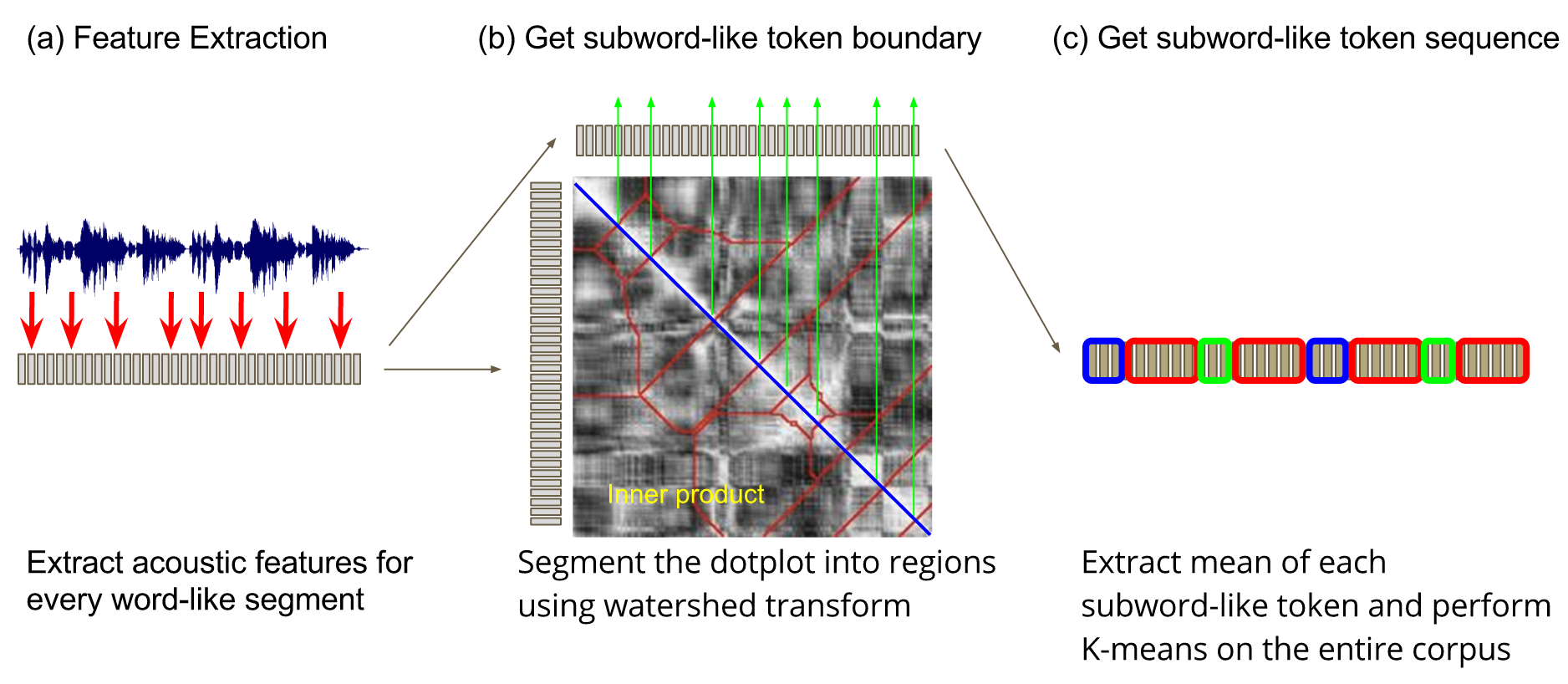}
     \caption{
An example dotplot of a word-like segment and its watershed transform in the initialization of the label set $W_0$.}
     \label{fig:WS}
\end{figure}

We obtain the initial labels in (\ref{eq:1}) in a top-down fashion by first breaking each utterance into word-like segments based on the discontinuities in a parameter evaluated from energy and MFCC features.
For each word-like segment, we further divide it into subword-like segments in the following way.
(a) We extract the acoustic features and (b) perform a watershed transform on the filtered self-similarity dotplot \cite{jansen2010towards} on the acoustic features of each hypothesized word-like segment.
Watershed transformation is able to capture the regions and their borders in a gray scale image \cite{couprie1997topological}.
So, the intersections of the diagonal entries of the dotplot with the region borders are taken as the boundaries between subword-like segments.
An example dotplot and its watershed transform including the hypothesized subword-like segment boundaries is shown in Fig. \ref{fig:WS}. 
(c) We then extract an average representative feature vector for every hypothesized subword-like segment, and perform global k-means clustering on these representative vectors obtained from the whole corpus.
%
%
A subword-like token ID is then assigned to each cluster.
A distinct sequence of consecutive subword-like tokens for word-like segments then defines a word-like token.
In this work we are only interested in the subword-like token sequence so for every utterance the sequence of word-like tokens are further flattened into a subword-like token sequence.
The corpus is thus represented by its initial subword-like token sequence $W_0$.

\subsubsection{Granularity Space of Multi-granular Acoustic Token Sets}\label{sec:2-2-2}
\begin{figure}[tb]
\centerline{\includegraphics[width=0.8\columnwidth]{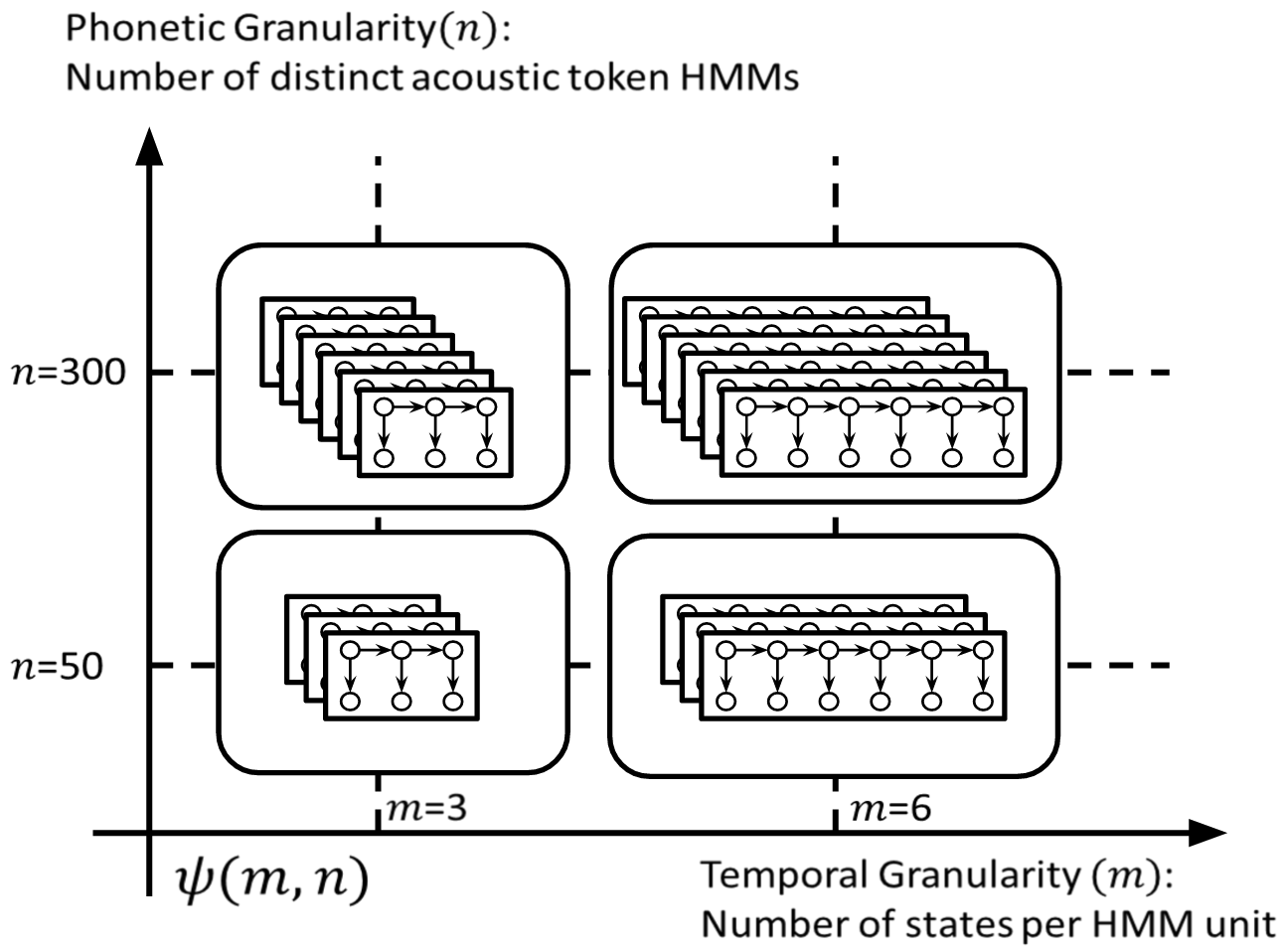}}
\caption{Model granularity space for HMM configurations}\label{fig:2dcube}
\end{figure}

The process explained above can be performed with different HMM configurations, each characterized by the hyperparameters $\psi=(m,n)$: the number of states $m$ in each acoustic token HMM, and the total number of distinct acoustic tokens $n$ during initialization. 
The token labels of an utterance can be considered as a temporal segmentation of the signal, so the HMM length (or number of states in each HMM) $m$ represents the temporal granularity. 
The set of all distinct acoustic tokens can be considered as a segmentation of the phonetic space, so the total number $n$ of distinct acoustic tokens represents the phonetic granularity. 
This gives a two-dimensional representation of the acoustic token sets in terms of temporal and phonetic granularities as in Fig. \ref{fig:2dcube}. 
Different points in this two-dimensional space in Fig. \ref{fig:2dcube} correspond to acoustic token configurations with different model granularities, carrying complementary knowledge about the corpus and the language. 
Although the best selection of the hyperparameters in the above two-dimensional space in general is not known, we can simply select $M$ temporal granularities ($m=m_1, m_2, ..., m_M$) and $N$ phonetic granularities ($n=n_1, n_2, ..., n_N$), forming a two-dimensional array of $M \times N$ hyperparameter pairs in the granularity space.

\subsection{Mutual Reinforcement(MR) of Multi-granular Tokens}
%
%
%

Mutual Reinforcement (MR) is intended as a replacement of the initialization step when multiple token sets are present. 
After given the initialization in Eq. \eqref{eq:1}, each acoustic token set for a different granularity (m,n) is trained independently with Eq. \eqref{eq:2}, and Eq. \eqref{eq:3}. We called this the Multi-granular Acoustic Tokenizer (MAT). 
Because each acoustic token set is trained independently without any ground-truth label, they may be susceptible to small variations in the acoustic features.
Mutual Reinforcement intends to remove such errors from individual token sets by comparing and fusing the information obtained from different token sets and serve as a replacement for initialization in Eq. \eqref{eq:1}. 
Once Mutual Reinforcement is complete, each acoustic token set for a different $(m,n)$ setting is trained independently with Eq. \eqref{eq:2} and Eq. \eqref{eq:3}, as shown in Fig. \ref{fig:3} (a) where we draw an arrow back to the MAT.
We now explain the details in Fig. \ref{fig:3}, including token boundary fusion and LDA-based token label re-initialization as in Fig. \ref{fig:3} (a).

\subsubsection{Token Boundary Fusion}
Fig. \ref{fig:3} (b) shows the token boundaries when a part of an utterance is segmented into acoustic tokens on different levels with different hyperparameter pairs $\psi=(m,n)$. 
We define a boundary function $b_{m,n} (j)$ on each level with $\psi=(m,n)$ for the possible boundary between every pair of two adjacent frames within the utterance, where $j$ is the time index for such possible boundaries.
On each level $b_{m,n} (j)=1$ if boundary $j$ is a token boundary and 0 otherwise.
All these boundary functions $b_{m,n} (j)$ for all different levels are then weighted and averaged to give a joint boundary function $B(j)$.
The weights consider the fact that smaller $m$ or shorter HMMs generate more boundaries, so those boundaries should weight less.
The peaks of $B(j)$ are then selected when the second derivative of $B(j)$ is below a threshold set empirically.
This gives the new segmentation of the utterance as shown at the bottom of Fig. \ref{fig:3} (b).

\subsubsection{LDA-based Token Label Re-initialization}
\begin{figure}[tbh]
\centerline{\includegraphics[width=0.96\columnwidth]{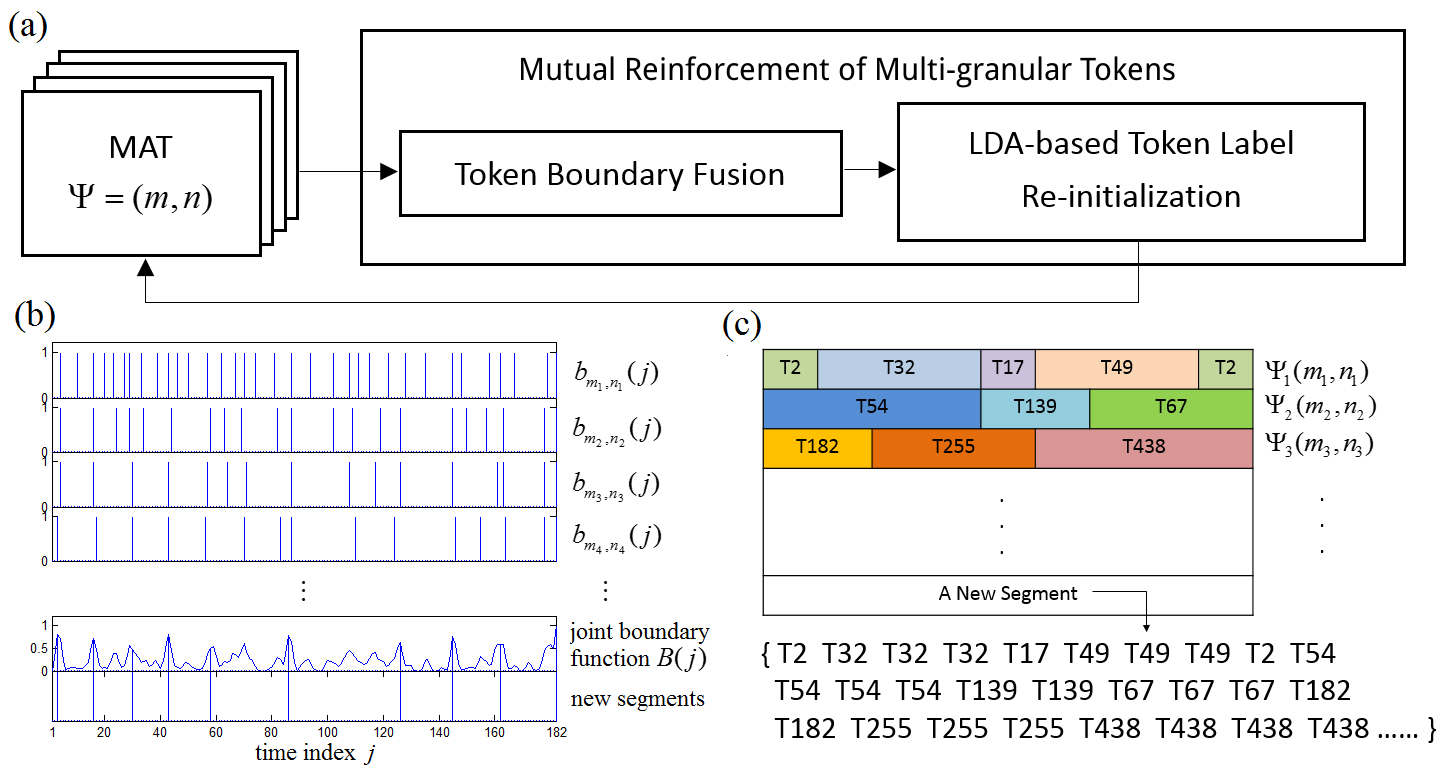}}
\caption{Mutual Reinforcement(MR) of multi-granular tokens: (a) block diagram, (b) token boundary fusion, and (c) a new segment considered as a document (bag-of-``pseudo-words'') and a token as a ``pseudo-word'' in LDA-based token label re-initialization.
}\label{fig:3}
\end{figure}

As shown in Fig. \ref{fig:3} (c), each new segment of signals obtained above (between two new boundaries found by $B(j)$) usually consists of a sequence of acoustic tokens on each level based on the tokens trained on that level. 
{\color{black}
We now consider all the tokens on all the different levels as different ``pseudo-words'', so we have a vocabulary of $M \sum_{i=1}^{N}     n_i$ ``pseudo-words'', i.e., there are $n_i$ ``pseudo-words'' for the $i$-th phonetic granularity and there are a total of $M$ temporal granularities.
}
A new segment here is thus considered as a document (bag-of-``pseudo-words'') composed of ``pseudo-words'' (tokens) collected from all different levels.
Latent Dirichlet Allocation \cite{blei2003latent} (LDA) can then be performed for topic modeling, and then each document (new segment) is labeled with the most probable topic.
A sequence of such segments thus becomes a sequence of IDs of the most probable topic, which we treat as the new sequence of tokens.
Because in LDA a topic is characterized by a word distribution, here a topic or token distribution across different levels obtained with LDA may also represent an acoustic characteristic, or a new acoustic token.
By setting the number of topics in LDA as the number of distinct tokens $n$ ($n = n_1, n_2, ..., n_N$) in the hyperparameters $\psi=(m,n)$ as in subsection \ref{sec:2-2-2}, we have a new set of $n$ tokens obtained for each $n$.
Each of these new tokens corresponds to an LDA topic.
%
This new set of tokens thus gives a new set of initial labels $W_0$ as in (\ref{eq:1}) of subsection \ref{sec:2-2-1}.
{\color{black}
This process is repeated for the number of topics being $n_1, n_2, ..., n_N$.
These $N$ segmentations become the new initial label in Eq (\ref{eq:1}) of the $M\times N$ sets of relabeled acoustic tokens. All $M$ acoustic token sets with the same phonetic granularity share the same initialization. The $M$ temporal granularities for each phonetic granularity $n_i$ slowly diverge in the training of Eqs (\ref{eq:2}), (\ref{eq:3}).
}

\subsection{The Multi-target DNN (MDNN)}
As shown in the right part of Fig. \ref{fig:1}, token label sequence from a level (with hyperparameters $\psi=(m,n)$) is a valid target for supervised framewise training, although obtained in an unsupervised way.
In the initial work here, we simply take the token label rather than the HMM state as the training target. 
{\color{black}
As shown in Fig. \ref{fig:1}, there are $M\times N$ sets of multi-granular token labels with different hyperparameters $\psi=(m,n)$ for each utterance offered by MAT on the left, so we jointly consider all $M\times N$ sets of multi-granular token labels by learning the parameters for a single DNN with a uniformly weighted cross-entropy objective at the output layer. 
}
As a result, the bottleneck feature (BNF) extracted from this DNN automatically fuse all knowledge about the given corpus and the language behind from the different sets of acoustic tokens.


\subsection{Iterative Learning with MAT-DNN}
\label{sec:2-5}
Once the bottleneck features (BNFs) are extracted from the MDNN in iteration 1, they can be taken as the input to the MAT on the left of Fig. \ref{fig:1} replacing the initial acoustic features. 
The MAT then generates updated sets of multi-granular token labels and these updated sets of multi-granular token labels can be used as the updated training targets of the MDNN. 
The input features of the MDNN can also be updated by concatenating the initial acoustic features with the newly extracted BNFs as the concatenated features.
This iterative process can be repeated.
%
The concatenated feature used as the input of the MDNN can be further augmented by concatenating unsupervised features obtained with other approaches such as the Deep Boltzmann Machine \cite{salakhutdinov2009deep} (DBM) posteriorgrams, Long Short-Term Memory Recurrent Neural Network \cite{hochreiter1997long} (LSTM-RNN) autoencoder bottleneck features, and i-vectors \cite{kanagasundaram2011vector} trained on MFCC.
{\color{black}
Although different from the conventional recurrent neural network (RNN) in which the recurrent structure is included in the backpropagation training, the concatenation of the bottleneck features with other features in the next iteration in MDNN is a kind of recurrent structure where the information propagates between the MAT and the MDNN.
}

\subsection{Query-by-Example Spoken Term Detection}
\label{sec:2-6}
Spoken Term Detection usually refers to the task of finding all occurrences of the test query from a target audio corpus.
When the query is also spoken, the task is referred to as query-by-example spoken term detection.
With the multi-granular acoustic tokens obtained above, we have many sets of tokens each defined by a hyperparameter set $\psi=(m,n)$, and the query-by-example spoken term detection can be performed as follows.

Let \{$p_r, r=1,2,3,..,n$\} denote the $n$ acoustic tokens in the set defined by $\psi$=$(m,n)$.
We first construct a distance matrix $S$ of size $n \times n$ off-line for every token set $\psi=(m,n)$, for which the element $S(i,j)$ is the distance between any two token HMMs $p_i$ and $p_j$ in the set,
\begin{equation}
S(i, j) =\mbox{KL} (i, j). \label{eq:soft}
\end{equation}
The KL-divergence $\mbox{KL} (i,j)$ between two token HMMs in (\ref{eq:soft}) is defined as the symmetric KL-divergence between the states based on the variational approximation \cite{hershey2007approximating} summed over the states. 

\begin{figure}[b]
\centerline{\includegraphics[width=0.3\textwidth]{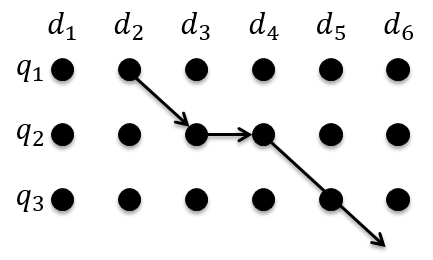}}
\caption{Token-level DTW over the matching matrix $W^T$ for a simplified  example with $D=6$ and $Q=3$.}\label{fig:dtw}
\end{figure}

In the on-line phase, we perform the following for each entered spoken query $q$ and each document (utterance) $d$ in the target corpus for each token set $\psi=(m,n)$. 
Assume for a given token set a document $d$ is decoded into a sequence of $D$ acoustic tokens with indices $(d_1, d_2, ..., d_D)$ and the query $q$ into a sequence of $Q$ tokens with indices $(q_1, ..., q_Q)$. 
We thus construct a matching matrix $W$ of size $D \times Q$ for every document-query pair, in which each entry $(i,j)$ is the distance between acoustic tokens with indices $d_i$ and $q_j$ as in (\ref{eq:topk}), where $S(i,j)$ is defined in (\ref{eq:soft}),
\begin{equation}
W(i, j)  = S(d_i, q_j).  \label{eq:topk}
\end{equation}
A simplified example for $D=6$ and $Q=3$ is shown in Fig. \ref{fig:dtw}.
We then perform token-level DTW on this matching matrix $W$ by summing the distance between token pairs along the optimal path and return the minimal distance as the distance between document (utterance) $d$ and query $q$ on the considered level of tokens with hyperparameter $\psi=(m,n)$.
Because it is not known which level with $\psi=(m,n)$ is more useful, we simply sum the distances over all the $M\times N$ levels, and the knowledge collected from tokens on different levels will be complementary to each other.
This sum is used to rank the documents (utterances) for the query.

Note that in this approach we are simply matching the token sequence in $q$ with that in an utterance $d$ without knowing exactly which words are spoken in $q$ and $d$. 
%
%
The matching is performed completely on the token level rather on the word level, so supervised ASR is not needed.
Also matching on the token level requires much less computation than on the frame level.

\section{Experimental Setup}\label{sec:III}
\subsection{Setup of MAT-DNN}

The MAT-DNN framework presented above allows flexible configurations, but here in the preliminary experiments we trained the MAT-DNN in the following manner. 
We set $m=3, 5, 7, 9$ states per token HMM and $n=50, 100, 300, 500$ distinct tokens in the MAT, which gave a total of 16 levels ($m=11,13$ were added in some tests as mentioned below).

In the first iteration, we used the 39 dimension MFCC with energy, delta and double delta as the initial acoustic features for the input to both the MAT and the MDNN.
We concatenated 4 frames before and after (39x9 dimensions), and an i-vector (400 dimensions) trained on the MFCCs as the input of the MDNN. 
The topology of the MDNN was set to be 751(input)-256(hidden)-256(hidden)-39(bottleneck)-(target) with 3 hidden layers. 
%
%
We kept the dimensionality of the bottleneck features and all considered features to be 39 for a fair comparison. 
{\color{black}
For the Deep Boltzmann Machine(DBM)\cite{zhang2012resource}, we used the 39-dimension MFCCs with a window of 5 frames before and after as the input. 
}
The configuration we used for the DBM was 429(visible)-256(hidden)-256(hidden)-39(hidden). 
We also extracted another set of LSTM-RNN autoencoder bottleneck features but found the performance was slightly worse than MFCCs.

In the second iteration, we concatenated the original MFCCs, the bottleneck features (BNF) extracted from the first iteration, and the DBM posteriorgrams, all with 4 frames before and after, and the i-vector forming a 1453 ($39\times 9+39\times 9+39\times 9+400$) dimension input to the MDNN. 
We used the updated token labels as the target and extracted the BNF as the features. 
{\color{black}
The two corpora used in the Zero Resource Speech Challenge were used here for easier comparison of results: the Buckeye corpus \cite{pitt2007buckeye} (14137 utterances) in English and the NCHLT Xitsonga Speech corpus (4058 utterances) in Tsonga.
}
The MAT was trained using the zrst \cite{chung2014zero}, a python wrapper for the HTK toolkit \cite{young1997htk} and srilm \cite{stolcke2002srilm} for training unsupervised HMMs with varying model granularity.
%
%
%
The i-vectors were extracted using Kaldi \cite{Povey_ASRU2011}. 
%
%
The MDNN was trained using  Caffe \cite{jia2014caffe}.

\subsection{Evaluation Metrics for Discovered Features and Tokens}\label{sec:IV}
The evaluation metrics defined by the Zero Resource Speech Challenge of Interspeech 2015 \cite{versteegh2015zero} were used in this work.
For evaluating the discovered frame-level features, the minimum pair ABX task \cite{schatz2013evaluating,schatz2014evaluating} was used to measure the discriminability between two sound categories as in Track 1 of the Challenge. 
For evaluating the discovered tokens in Track 2 of the Challenge, a total of seven evaluation metrics were defined:
Normalized Edit Distance (NED), Coverage, Matching F-score, Grouping F-score, Type F-score, Token F-score and Boundary F-score \cite{versteegh2015zero}.


\subsection{Evaluation of Query-by-Example Spoken Term Detection}\label{sec:4c}
We further tested the proposed approach with a real application: query-by-example spoken term detection.
In this task, the user gives a spoken query, and the system is supposed to return a ranked list of spoken documents that contain the query.
The documents in the list are ranked by the relevance scores evaluated between the spoken query and each of the document (utterance) in the target corpus.
This returned document list is then compared to the ground truth document list using the standard metric for retrieval: Mean Average Precision (MAP) \cite{philbin2007object,yue2007support}. 
This task of spoken term detection allows both the obtained frame-level features and the discovered tokens to be evaluated in a single framework.
For frame-level features, we can perform the frame-level DTW \cite{muller2007dynamic} between sequences of features obtained by the MAT-DNN for the query and each spoken document; 
for the discovered tokens we can decode both the spoken query and each document in the target corpus into sequences of acoustic tokens, and then perform token-level DTW as explained in section \ref{sec:2-6}.

\section{Experimental Results}\label{sec:V}
Experimental results for the proposed approaches are reported in this section.

\subsection{Quality of the Frame-level Feature in Metrics of Track 1}
The evaluation was based on the ABX discriminability test \cite{schatz2013evaluating} 
%
defined for Track 1 of the Challenge \cite{versteegh2015zero}.
The results in error percentage (the lower the better) are listed in Table \ref{tab:1}.
The first two columns are for the English corpus while the next two columns are for the Tsonga corpus defined by the Challenge, both including across speaker and within speaker tasks.


\begin{figure*}[tb]
\centerline{\includegraphics[width=\textwidth]{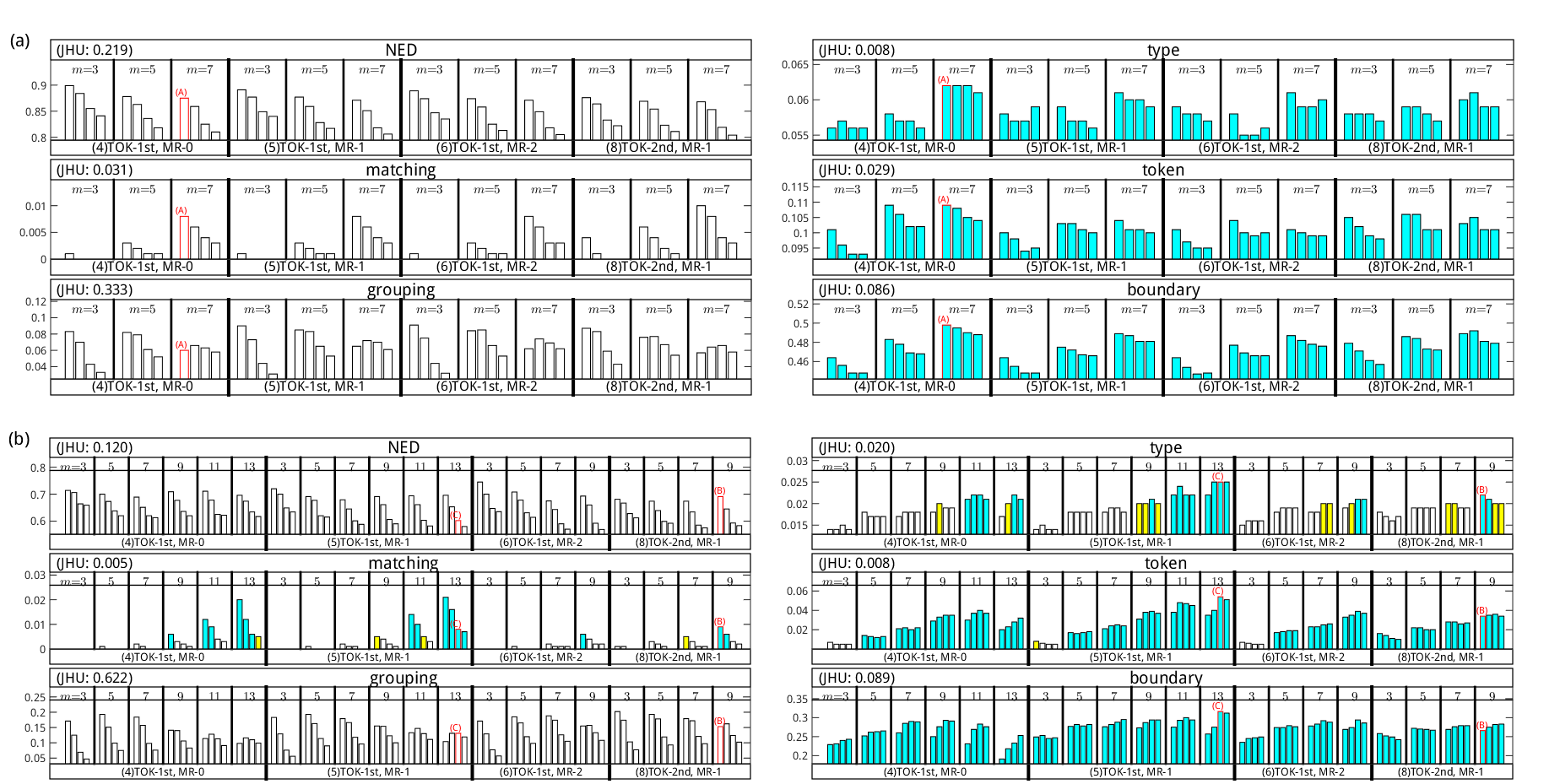}}
\caption{Quality of discovered acoustic tokens in metrics of Track 2 for (a) English and (b) Tsonga. 
Each subgraph is for an evaluation measure, including four sections from left to right for four cases of token sets 
as marked at the bottom.
The four bars in each group for a value of $m$ are for $n=50, 100, 300, 500$ from left to right (not shown in the figure) and $\psi=(m,n)$ are parameters for the token sets. Blue, yellow and white bars correspond to better, equal to or worse as compared to the JHU baseline offered by the Challenge listed at the upper left corner of each subgraph. 
}
\label{fig:t1}
\end{figure*}

\begin {table}[tb]

\caption {Frame-level Speech feature quality in the metrics (ABX error percentages) of Track 1 of the Challenge for the languages English and Tsonga, both including across speaker and within speaker tasks. 
The best figure for each metric is shown in bold. \label{tab:1}}
\begin{center}
\resizebox{0.95\columnwidth}{!}{%
\begin{tabular}{|c|l|r|r|r|r|r|}
\hline
\multicolumn{2}{|c|}{\multirow{2}{*}{Method}} & \multicolumn{1}{c|}{\multirow{2}{*}{dim}} & \multicolumn{2}{c|}{English}                              & \multicolumn{2}{c|}{Tsonga}                               \\ \cline{4-7} 
\multicolumn{2}{|c|}{}                        & \multicolumn{1}{c|}{}                     & \multicolumn{1}{c|}{across} & \multicolumn{1}{c|}{within} & \multicolumn{1}{c|}{across} & \multicolumn{1}{c|}{within} \\ \hline
(1)   & Baseline           &13    & 28.1    & 15.6    & 33.8     & 19.1          \\ \hhline{|=|=|=|=|=|=|=|}
(2)   & MFCC               &39 & 28.6    & 15.9    & 30.8     & 16.3          \\ \hline
(3)   & DBM posterior &39    & 26.0    & 15.7    & 29.2     & 16.2          \\ \hline
(4)   & BNF-1st, MR-0   &39   & 26.8    & 16.0    & 26.5     & 15.5          \\ \hline
(5)   & BNF-1st, MR-1     &39  & 23.9    & 14.6    & 22.0     & 13.4          \\ \hline
(6)   & BNF-1st, MR-2     &39  & 24.5    & 14.9    & 22.1     & 13.3          \\ \hline
(7)   & BNF-2nd, MR-0    &39   & 26.6    & 16.3    & 26.2     & 15.1          \\ \hline
(8)   & BNF-2nd, MR-1    &39   & 24.5    & 15.1    & 23.3     & 13.9          \\ \hline
(9)  & BNF-2nd, MR-1*    &39  & 23.0    & 15.1    & 22.7     & 15.2          \\ \hline
(10)   & BNF-2nd, MR-2*   &39  & 23.2    & 14.7    & 23.3     & 14.3          \\ \hline
(11)   & BNF-1st, MR-1,64  &64  & 23.2    & 14.5    & 21.9     & 13.3          \\ \hline 
(12)  & BNF-1st, MR-1,128  &128 & 23.0    & 14.5    & 21.9     & 13.3          \\ \hline
(13)  & BNF-1st, MR-1,256  &256 & \textbf{21.9}  & \textbf{14.0} & \textbf{21.4} & \textbf{12.8} \\ \hline
(14)   & LSTM autoencoder  &50  & 36.1    & 23.9   & 34.3   & 20.5           \\ \hline
(15)   & MAT-LSTM  &39  & 23.2 & 14.8 & 24.0 & 14.7          \\ \hhline{|=|=|=|=|=|=|=|}
(16)   & Thiolli\`ere et al. \cite{thiolliere2015hybrid}    &100  & 17.9    & 12.0    & 16.6      & 11.7         \\ \hline
(17)   & Renshaw et al. \cite{renshaw2015comparison}  &100    & 21.1    & 13.5    & 19.3      & 11.9         \\ \hline
(18)   & Badino et al. \cite{badino2015discovering}    &64   & 26.3    & 17.3    & 23.6      & 14.1         \\ \hline
(19)   & Chen et al. \cite{chen2015parallel}    &  385 & 16.3    & 10.8    & 17.2      & 9.6         \\ \hline
(20)   & Baljekar et al. \cite{baljekar2015using} & 26     & 29.8    & 18.4    & 29.7      & 18.1         \\ \hhline{|=|=|=|=|=|=|=|}
(21)  & Topline       &      & 16.0    & 12.1    & 4.5     & 3.5          \\ \hline
\end{tabular}
}
\end{center}
\end{table}

Rows (1) and (21) are the official baseline (MFCC features without delta and double delta) and the official topline (supervised phone posteriorgrams) provided by the Challenge respectively. 
Row (2) is our baseline of the MFCC features with delta and double delta, the initial acoustic features used to train all systems in this work. 
We can see our baseline is very close to the baseline provided by the Challenge.
Row (3) is for the Deep Boltzmann Machine (DBM) posteriorgrams extracted from the MFCC of row (2), serving as a strong unsupervised baseline, which is clearly better than row (2) in all cases. 
The results in rows (4), (5) and (6) are for the bottleneck features extracted in the first iteration (BNF-1st) of the MAT-DNN without applying Mutual Reinforcement (MR) (MR-0, row (4)), applying MR once (MR-1, row (5)), and twice (MR-2, row (6)) respectively.
Rows (7) and (8) are trained in a similar way to rows (4) and (5) with MR not applied and applied once, except the bottleneck features were extracted in the second iteration of the MAT-DNN (BNF-2nd) and the MAT was trained using the BNF of row (5).
Row (9) is similar to row (8), except in the input of the MDNN only concatenation of the BNF of row (5) and the i-vectors was used.
Row (10) is similar to row (9), except the training targets of the MDNN were those obtained with two iterations of MR.
%
Row (11), (12) and (13) is similar to row (5), except we use a wider bottleneck layer with 64, 128, 256 dimensions instead of 39.
Row (14) is the performance of the 50 dimension bottleneck feature extracted from a (39$\times$9 input)-(50 hidden)-(50 bottleneck)-(50 hidden)-(39$\times$9 target) LSTM-RNN autoencoder.
%
Row (15) is the performance of the 39 dimension bottleneck feature extracted from a (39+400 input)-(256 hidden)-(256 hidden)-(39 bottleneck)-(Multi-target) LSTM-RNN classifier, where the ``Multi-target'' uses the same mult-granular acoustic token set in row (5).
%
Row (16) to (20) are the performance of the systems for by participants of the Challenge \cite{versteegh2015zero}.
%


All the features from row (2) to (10) are confined to 39 dimensions for fair comparison. 
We observe that as a stand-alone feature extractor without any iterations, the MAT-DNN in row (5) outperformed the DBM baseline in row (3) significantly.
The effect of Mutual Reinforcement can be seen in the improvement from rows (4) to (5), (6) and rows (7) to (8). 
We observe that a single iteration of Mutual Reinforcement is enough to bring huge improvement, while the second iteration did not improve as much. 
In other words, the Mutual Reinforcement is quite capable in fusing knowledge from all other levels of tokens, so a single iteration was enough to saturate the performance improvement that can gained from using information from other token sets.
The effect of iterations in the MAT-DNN can be seen by comparing rows (2), (5), (8), respectively corresponding to 0, 1, and 2 iterations. 
Although the performance improvement from row (2) to row (5) is notable, the performance dropped in the second iteration in (8). 
To investigate reasons of this performance drop, we widened the bottleneck features to 64, 128 and 256 dimensions in rows (11), (12), (13) and observed a dramatic improvement.
It is possible that we have not yet explored the full potential of the MAT-DNN.
For a better tuned set of parameters, improvement in the following iterations may be expected. 
%
{\color{black}
To answer the question how crucial the Multi-granular objective is to the feature extraction, in row (15) we replaced the DNN in Fig. \ref{fig:1} with a LSTM-RNN to predict the same multi-granular targets as in row (5). Although the performance is not as great as row (5), by comparing row (15) to (14), we get a conclusion that the Multi-granular objective contributes more to the performance than the network structure.}

When compared to the results in rows (16) to (20) from the participants of the Challenge, rows (15), (16), (18) seem to perform better, but the systems in rows (16), (17) and (19) had much larger feature dimension than rows (2) to (10). 
Although rows (11) to (13) have comparable dimensions and perform worse than (16), (17), and (19), their dimension was only increased in the last step of the MDNN in the bottleneck layer. 
%
%
The focus of this work is to propose a method to iteratively learn both the features and the tokens with a target application and to visualize the results.
In this work, we took the metrics of the Challenge as a reference point rather than an objective.
%
%
It is definitely interesting to tune the many parameters here, but this is beyond the scope of this work.
%

\subsection{Quality of the Acoustic Tokens in Metrics of Track 2}
\begin{table*}[tb]
\caption{Comparison of three typical example token sets (A)(B)(C) selected out of all shown in Fig. \ref{fig:t1} with the JHU baseline offered by the Challenge, the systems proposed by R\"as\"anen et al. \cite{rasanen2015unsupervised}, and Lyzinki et al. \cite{lyzinski2015evaluation} in terms Precision (P), Recall (R), and F-score (F). Those better than JHU baseline are in bold.}
\centering
\tabcolsep=0.11cm
\resizebox{0.95\textwidth}{!}{%
\begin{tabular}{|l|lr|rr|rrr|rrr|rrr|rrr|rrr|}
\hline
\multicolumn{1}{|c|}{\multirow{2}{*}{Lan}} & \multicolumn{2}{c|}{\multirow{2}{*}{System}} & \multicolumn{1}{c}{NED} & \multicolumn{1}{c|}{Cov} & \multicolumn{3}{c|}{Matching}                                          & \multicolumn{3}{c|}{Grouping}                                          & \multicolumn{3}{c|}{Type}                                              & \multicolumn{3}{c|}{Token}                                             & \multicolumn{3}{c|}{Boundary}                                          \\
\multicolumn{1}{|c|}{}                     & \multicolumn{2}{c|}{}                        &                         &                          & \multicolumn{1}{c}{P} & \multicolumn{1}{c}{R} & \multicolumn{1}{c|}{F} & \multicolumn{1}{c}{P} & \multicolumn{1}{c}{R} & \multicolumn{1}{c|}{F} & \multicolumn{1}{c}{P} & \multicolumn{1}{c}{R} & \multicolumn{1}{c|}{F} & \multicolumn{1}{c}{P} & \multicolumn{1}{c}{R} & \multicolumn{1}{c|}{F} & \multicolumn{1}{c}{P} & \multicolumn{1}{c}{R} & \multicolumn{1}{c|}{F} \\ \hline
                                           & (1) JHU Baseline               &             & 21.9                    & 16.3                     & 39.4                  & 1.6                   & 3.1                    & 21.4                  & 84.6                  & 33.3                   & 6.2                   & 1.9                   & 2.9                    & 5.5                   & 0.4                   & 8.0                    & 44.1                  & 4.7                   & 8.6                    \\ \cline{2-20} 
                                           & (2) Proposed                   & (A)         & 87.5                    & \textbf{100.0}           & 1.4                   & 0.5                   & 0.8                    & 3.6                   & 18.7                  & 6.0                    & 4.2                   & \textbf{11.9}         & \textbf{6.2}           & \textbf{8.3}          & \textbf{15.7}         & \textbf{10.9}          & 35.2                  & 84.6                  & 49.8                   \\ \cline{2-20} 
English                                    & (3) R\"as\"anen et al.         &             & 70.8                    & 42.4                     &                       &                       &                        & 13.4                  & 15.7                  & 14.2                   & 14.1                  & 12.9                  & 13.5                   & 22.6                  & 6.1                   & 9.6                    & 75.7                  & 33.7                  & 46.7                   \\ \cline{2-20} 
                                           & (4) Lyzinski et al.            &             & 61.2                    & 80.2                     & 6.5                   & 3.5                   & 4.6                    &                       &                       &                        & 3.1                   & 9.2                   & 4.6                    & 2.4                   & 3.5                   & 2.8                    & 18.8                  & 64.0                  & 29.0                   \\ \cline{2-20} 
                                           & (5) Topline                    &             & 0.0                     & 100.0                    & 98.3                  & 18.5                  & 31.1                   & 99.5                  & 100.0                 & 99.7                   & 50.3                  & 56.2                  & 53.1                   & 68.2                  & 60.8                  & 64.3                   & 88.4                  & 86.7                  & 87.5                   \\ \hline
                                           & (1) JHU Baseline               &             & 12.0                    & 16.2                     & 69.1                  & 0.3                   & 0.5                    & 52.1                  & 77.4                  & 62.2                   & 3.2                   & 1.4                   & 2.0                    & 2.6                   & 0.5                   & 0.8                    & 22.3                  & 5.6                   & 8.9                    \\ \cline{2-20} 
                                           & (2) Proposed                   & (B)         & 69.1                    & \textbf{95.0}            & 5.9                   & \textbf{0.5}          & \textbf{0.9}           & 10.7                  & 26.8                  & 15.3                   & 1.5                   & \textbf{3.9}          & \textbf{2.2}           & 2.3                   & \textbf{6.6}          & \textbf{3.4}           & 17.1                  & \textbf{59.1}         & \textbf{26.6}          \\
                                           &                                & (C)         & 60.2                    & \textbf{96.1}            & 9.7                   & \textbf{0.4}          & \textbf{0.8}           & 13.5                  & 12.7                  & 13.1                   & 1.8                   & \textbf{4.7}          & \textbf{2.5}           & \textbf{3.9}          & \textbf{9.1}          & \textbf{5.4}           & 21.2                  & \textbf{62.1}         & \textbf{31.6}          \\ \cline{2-20} 
Tsonga                                   & (3) R\"as\"anen et al.         &             & 36.4                    & 94.7                     &                       &                       &                        & 10.7                  & 3.3                   & 5.0                    & 2.2                   & 6.2                   & 3.3                    & 2.3                   & 3.4                   & 2.7                    & 29.2                  & 39.4                  & 33.5                   \\ \cline{2-20} 
                                           & (4) Lyzinski et al.            &             & 43.2                    & 89.4                     & 21.2                  & 3.8                   & 6.5                    &                       &                       &                        & 4.9                   & 18.8                  & 7.8                    & 2.2                   & 12.6                  & 0.8                    & 18.8                  & 64.0                  & 29.0                   \\ \cline{2-20} 
                                           & (5) Topline                    &             & 0.0                     & 100.0                    & 100.0                 & 6.8                   & 12.7                   & 100.0                 & 100.0                 & 100.0                  & 15.1                  & 18.1                  & 16.5                   & 34.1                  & 49.7                  & 40.4                   & 66.6                  & 91.9                  & 77.2                   \\ \hline
\end{tabular}
}
\label{tab:2}
\end{table*}

Track 2 of the Challenge defined a total of seven evaluation metrics \cite{ludusan2014bridging} describing different aspects of the quality of the acoustic tokens discovered.
%
Except for Coverage and NED whose values are indicators of the system characteristic rather than the system performance, the higher the value the better for the other five metrics.
Except for Coverage, the other six scores are shown in the six subfigures in Fig. \ref{fig:t1} (a) for English and (b) for Tsonga. We omit Coverage here because with our approach Coverage is always 100\% in all cases.
In each subfigure, the results for four cases are shown in four sections from left to right, corresponding to the four sets of tokens obtained in MAT after the first and second iterations of MAT-DNN (marked by TOK-1st or TOK-2nd) with MR performed or not (MR-0,1,2). 
The corresponding bottleneck features for them are those listed in rows (4), (5), (6) and (8) of Table \ref{tab:1}.
These are marked at the bottom of each section.
For each of these sections, the three or six groups of bars correspond to different values of $m$ ($m=3, 5, 7$ or $m=3, 5, 7, 9, 11, 13$), while in each group the four bars correspond to the four values of $n$ ($n=50, 100, 300, 500$ from left to right), where $\psi=(m,n)$ are the parameters for the token sets. 
The bars in blue and yellow are those better or equal to the JHU baseline \cite{jansen2011efficient} offered by the Challenge, while those in white are worse.
%
%
Only the results jointly considering both within and across talker conditions are shown.

From Fig. \ref{fig:t1} (a) for English, it can be seen that the proposed token sets performed well in Type, Token and Boundary F-scores, although much worse in Matching and Grouping F-scores. 
We see in many cases the benefits brought by MR (e.g. (6) TOK-1st, MR-2 vs (5) TOK-1st, MR-1 in Type F-score of Fig. \ref{fig:t1} (a)) and the second iteration (e.g. (8) TOK-2nd vs (5) TOK-1st in Boundary F-score of Fig. \ref{fig:t1} (a)), especially for small values of $m$.
In many groups for a given $m$, smaller values of $n$ seemed better, probably because $n=50$ is close to the total number of phonemes in the language.
Also, a general trend is that larger values of $m$ were better, probably because HMMs with more states were better in modeling the relatively longer linguistic units; this may directly lead to the higher Type, Token and Boundary F-scores.

Similar observations can be made for Tsonga in Fig. \ref{fig:t1} (b), in which additional results for $m=9, 11, 13$ are reported. 
We see the overall performance seemed to be even better as the proposed token sets performed well even in Matching F-scores. 
The improvements brought by MR (e.g. MR-1 vs MR-0), the bottleneck features (compared to JHU baseline) and the second iteration (TOK-2nd vs TOK-1st) are better observed here, which gave the best cases for all the five main scores. 
This is probably due to the fact that more sets of tokens were available for MR and MAT-DNN  on Tsonga than English. 
We can conclude from this observation that more token sets introduced more robustness and that led to better token sets for the next iteration.  
When $m$ went up to 13, we see that for (4) TOK-1st, MR-0 in the left section of Fig. \ref{fig:t1} (b) almost all metrics degraded except for Matching F-scores, but with MR-1, MR-2 almost all the F-scores consistently increased (except for NED) when $m$ became larger. 
This suggests that MR can prevent degradation from happening while detecting relatively longer linguistic units.

In Table \ref{tab:2}, we selected three typical example token sets (A), (B), and (C) out of the many proposed here in row (2), and compared them with the JHU baseline \cite{jansen2011efficient} in row (1), along with the representative results of the Challenge systems proposed by R\"as\"anen et al. \cite{rasanen2015unsupervised} in row (3), Lyzinski et al. \cite{lyzinski2015evaluation} in row (4), and the supervised topline system \cite{johnson2006adaptor} in row (5) in terms of the seven metrics including Precision (P), Recall (R) and F-scores (F). 
%
The three selected tokens sets are: (A): (TOK-1st, MR-0, $m=7$, $n=50$) for English; (B): (TOK-2nd, MR-1, $m=9$, $n=50$) and (C): (TOK-1st, MR-1, $m=13$, $n=300$) for Tsonga.
These three selected proposed example sets are also marked in Fig. \ref{fig:t1}. 
%
%
%

We first compare the proposed sets (A), (B), and (C) with the JHU baseline.
%
%
%
In Table \ref{tab:2} those of the proposed approaches (A), (B), and (C) better than JHU baseline are in bold.
Regarding the NED and coverage, a better system should have lower NED and higher coverage. 
A system that discovers a lot of tokens usually has high coverage and also high NED, because more tokens usually means more mismatches. 
Such a system is said to be permissive. 
On the other hand a system that only returns high confidence tokens usually has low coverage and also low NED. 
Such a system is said to be selective.
Therefore NED and coverage are closely related and have to do with the system characteristics and engineering trade-offs.
As a result, the much higher NED and coverage scores of the proposed token sets (A), (B), and (C) suggest that the proposed approach is highly permissive, while the JHU baseline is highly selective.
The much higher parsing scores (Type, Token and Boundary scores), especially the Recall and F-scores, imply the proposed approach is more successful in discovering word-like units. 
However, the Matching and Grouping F-scores of sets (A), (B), and (C) were much worse, probably because the discovered tokens covered almost the whole corpus, including short pauses or silence, and therefore many tokens were actually noises. 
Another possible reason might be that the values of $n$ used were much smaller than the size of the real word vocabulary, making the same token label used for signal segments of varying characteristics, and this degraded the grouping quality.

When compared to the representative results of the Challenge in row (3) for English, the NED of our proposed token set (A) is
comparable to those of R\"as\"anen et al. \cite{rasanen2015unsupervised} while having significantly higher Coverage. 
It did better in terms of Token and Boundary F-scores while losing on Grouping and Type F-scores. 
For Tsonga, our token sets (B), (C) did not do better for NED and Coverage but did much better on Grouping and Token F-score with comparable Type and Boundary F-scores.
When compared to the representative results of the Challenge in rows (4) 
we see our approach was more permissive, 
%
%
doing much better on token F-score, and having similar performance on Boundary F-score on both languages.

To summarize the comparisons, the proposed MAT-DNN is a permissive approach that is very competitive in Token and Boundary F-scores.

\subsection{Unsupervised Spoken Term Detection}

\begin{figure*}[th!]
\centerline{\includegraphics[width=0.9\textwidth]{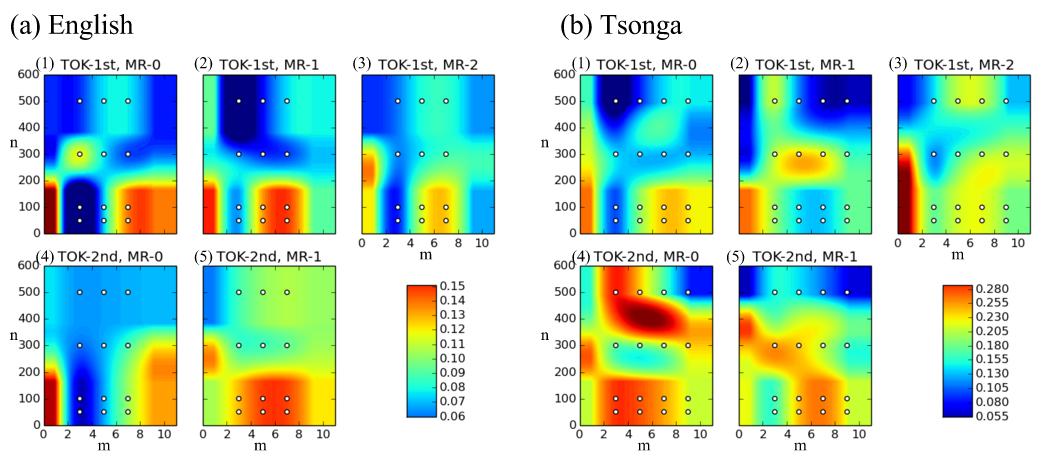}}
\caption{Performance distributions for Spoken Term Detection over the granularity space $(m,n)$ for (a) English and (b) Tsonga. 
Every white dot represents a chosen granularity $\psi=(m,n)$. 
The values for points around are those interpolated with 2D-spline.
This figure shows the performance of the individual token set with $\psi=(m,n)$ before they were averaged in Table \ref{tab:x}.
The details of the corresponding distribution in each image is in Table \ref{tab:heat}.
}
\label{fig:heat}
\end{figure*}

%
Separate experiments for query-by-example spoken term detection were conducted on the two corpora used here, English and Tsonga. 
For English and Tsonga, spoken instances of 5 and 10 query words randomly selected from the data set were used as the spoken queries to search for other instances in the spoken archive respectively. 
%
%
Both the selected queries and the corpora were first decoded as sequences of the multi-granular tokens. 
The distance between the document token sequences and query token sequence was evaluated by the token-level DTW distance over the matching matrix as defined in section \ref{sec:2-6} as opposed to conventional frame-level DTW. 

A total of 5 collections of multi-granular token sets were tested here, which are  (TOK-1st, MR-0), (TOK-1st, MR-1), (TOK-1st, MR-2), (TOK-2nd, MR-0), (TOK-2nd, MR-1). 
For English, each collection consists of 3$\times$4 sets of acoustic tokens with granularity $m = 3, 5, 7$ and $n = 50, 100, 300, 500$, so we obtained 12 scores for every query-document pair for every collection. 
For Tsonga, $m = 3, 5, 7, 9$ and $n = 50, 100, 300, 500$, thus we had 16 scores for every query-document pair for every collection. 
Mean Average Precision (MAP), the higher the better, was used as our evaluation metric.

The MAP performance for each of the five collections of token sets over the granularity plane $(m,n)$ is shown in Fig. \ref{fig:heat} (a) and (b) for English and Tsonga respectively.
The white points are the specific granularities chosen in the experiments, $\psi=(m,n)$, while the values around the white points are interpolated with 2D-spline.
The average, standard deviation, maximum, and minimum values evaluated over the 12 or 16 MAP values for each token collection in Fig. \ref{fig:heat} are further depicted in Table \ref{tab:heat}. 
%
%
It is clear from Fig. \ref{fig:heat} that the performance varied over the $(m,n)$ plane across different tokens collections.
It may be possible that the different high performing regions in Fig. \ref{fig:heat} have to do with the difference in the underlying linguistic structures of the corpus.
For example, for the token collection TOK-1st, MR-1, the performance distributions in Fig. \ref{fig:heat} (a2) and (b2) for English and Tsonga looked very different.
In Fig. \ref{fig:heat} (a2) for English there is a high-performing region near 50 tokens and 7 states per token; while in Fig. \ref{fig:heat} (b2) the high-performing region of Tsonga is closer to 300 tokens and 5 states per token.
We can observe a similar situation with the token collection (TOK-2nd, MR-0) in Fig. \ref{fig:heat} (a4) and (b4).
On the other hand, for some other token collections the performance distributions looked similar across the two languages.
For example, the token collection (TOK-1st, MR-0), Fig. \ref{fig:heat} (a1) and (b1) looked similar.
For the token collection (3) (TOK-1st, MR-2), Fig. \ref{fig:heat} (a3) and (b3) also looked similar to some degree.

The effect of MR can be observed by comparing rows (a1) to (a2),  (a2) to (a3), (a4) to (a5) for English in Table \ref{tab:heat}. 
Generally speaking, MR successfully reduced the variance between the performance of the individual acoustic token sets with different granularities $\psi=(m,n)$ and improved the overall and average performance.
Such improvements are also observable in Fig. \ref{fig:heat} by comparing (a1) to (a2) and (a4) to (a5).
Although this trend is less obvious for Tsonga in Table \ref{tab:heat}, from the performance distributions in Fig. \ref{fig:heat} (b2) to (b3), (b4) to (b5), we see the high performing regions more or less got smeared more evenly on the (m,n) plane.
The effect of the second iteration of the MAT-DNN can be observed by comparing rows (a1) to (a4), (a2) to (a5), (b1) to (b4), (b2) to (b5) in Table \ref{tab:heat} as well as the corresponding performance distributions in Fig. \ref{fig:heat}.
The second iteration almost consistently offered better performance than the first iteration.
A closer comparison of the performance distributions (a1) to (a4) and (b1) to (b4), (b2) to (b5) of Fig. \ref{fig:heat} reveals that although the performance was improved, the performance distributions produced by the second iteration look roughly similar to that produced by the first iteration for English in Fig. \ref{fig:heat} (a), although not as obvious for Tsonga in Fig. \ref{fig:heat} (b).


%
\begin{table}[tb]
\caption {The average, standard deviation, maximum and minimum values of the mean average precision for the spoken term detection experiments for every acoustic token collection in Fig. \ref{fig:heat}. \label{tab:heat}}
\begin{center}
\begin{tabular}{|l|l|l|r|r|r|r|}
\hline
\multirow{2}{*}{Lan.} & \multicolumn{2}{c|}{\multirow{2}{*}{token collections}} & \multicolumn{4}{c|}{MAP(\%)} \\ \cline{4-7} & \multicolumn{2}{c|}{} &avg   & std   & max   & min  \\ \hline
\multirow{5}{*}{Eng.}& (a1)& TOK-1st, MR-0& 7.78& 4.02& 17.06& 2.55   \\ \cline{2-7} 
                     & (a2)& TOK-1st, MR-1& 8.44& 3.39& 15.21& 2.50   \\ \cline{2-7} 
                     & (a3)& TOK-1st, MR-2& 8.43& 2.67& 13.35& 4.90   \\ \cline{2-7} 
                     & (a4)& TOK-2nd, MR-0& 9.10& 3.09& 16.41& 6.67   \\ \cline{2-7} 
                     & (a5)& TOK-2nd, MR-1& 10.63& 2.31& 14.96& 6.10  \\ \hline
\multirow{5}{*}{Tso.}& (b1)& TOK-1st, MR-0& 16.16& 5.35& 26.99& 4.94  \\ \cline{2-7} 
                     & (b2)& TOK-1st, MR-1& 14.35& 6.47& 26.87& 4.17  \\ \cline{2-7} 
                     & (b3)& TOK-1st, MR-2& 18.61& 6.34& 32.80& 7.59  \\ \cline{2-7} 
                     & (b4)& TOK-2nd, MR-0& 21.92& 5.52& 32.65& 8.20  \\ \cline{2-7} 
                     & (b5)& TOK-2nd, MR-1& 18.77& 6.49& 28.10& 5.66  \\ \hline
\end{tabular}
\end{center}
\end{table}

%
%
%
The results for the 5 collections of tokens shown in Fig. \ref{fig:heat} when the 12 or 16 distances from each collection of tokens were averaged are in rows (1) to (5) in Table \ref{tab:x}. 
The benefit of the iterative framework of Mutual Reinforcement (MR) can be observed by comparing rows (1) to (2), (2) to (3) and (4) to (5) (MR-0 vs MR-1, MR-1 vs MR-2). 
The benefit of the iterative framework of the MAT-DNN can also be observed by comparing rows (1) to (4) and (2) to (5) (TOK-1st vs TOK-2nd). 
We then compared these results with frame-level DTW performed on two different frame-level features: first the conventional 39-dim MFCC or the baseline in row (6), and then the bottleneck features (BNF-1st, MR-1) in row (7). 
When comparing rows (1)-(5) to the baseline in row (6), we see the token-level DTW were much better than the baseline frame-level DTW over MFCC in most cases, in addition to the significantly lower on-line computation requirements.
By comparing rows (6) to (7), we observe that the features obtained by MAT-DNN performed significantly better than the MFCC from which they were derived.
We then averaged all token-level DTW distances in rows (1) to (5) in row (12), and obtained better results.
This shows that the information obtained in different collections were complementary to each other as well. 
We further fused the information from both the frame-level DTW and token-level DTW by averaging all scores in rows (1) to (7) in row (13), producing even better results indicating frame-level and token-level information are complementary.
Here in both rows (12) and (13), the fused scores are simply averaged without any weighting.

To provide a reference about the performance of supervised tokens and systems, we applied the supervised phoneme recognizers from the Brno University of Technology (BUT) speech group \cite{matejka2005phonotactic} for four languages: English, Hungarian, Czech, and Russian.
These were supervised phoneme recognizers trained with human labeled data.
In other words, we took the phonemes of these four languages as four sets of supervised acoustic tokens, and used them in the exact same way described in section \ref{sec:2-6}, i.e. transcribing the corpora and the queries in English and Tsonga in terms of these four sets of supervised tokens, and performing token-level DTW.
However, because the internal parameters of the HMM of the BUT recognizer were not transparent to the user, the KL-divergence distance $S(i,j)$ in Eq. \eqref{eq:soft} can not be evaluated. 
These distance values were therefore estimated in the following way.
Assume the Czech phonemes were used as supervised tokens, and the Tsonga was used as the corpus.
We used the Czech phoneme recognizer to transcribe the Tsonga corpus and queries used in this work into Czech phonemes. 
For each Czech phoneme (the token), the percentages of the realizations of this token whose central frames belonged to each of the ground truth Tsonga phonemes were used to construct a feature vector with dimentionality $L$, where $L$ being the total number of distinct phonemes in Tsonga, and the $l$-th component of the vector was the percentage for the $l$-th Tsonga phoneme.
In this way, the distance $S(i,j)$ in Eq. \eqref{eq:soft} of section \ref{sec:2-6} between two tokens $p_i$ and $p_j$ could then be estimated by the cosine distance between the feature vectors of $p_i$ and $p_j$.

The results for the supervised tokens obtained with BUT phoneme recognizers in English, Hungarian, Czech, and Russian as described above are respectively listed in rows (8), (9), (10) and (11) in Table \ref{tab:x}.
To speed up the decoding process, we used the distance obtained from frame-level DTW of MFCC in row (6) to filter out documents with large distances.
Out of the four sets of supervised tokens, BUT English in row (8) did by far the best on English because these tokens were trained on a matched language while much worse for supervised tokens trained on mismatched languages in rows (9), (10) and (11).
Note that the BUT English also performed better on Tsonga than the BUT Hungarian, Czech and Russian phoneme recognizers as well.
Tsonga was an unmatched language for all the 4 phoneme recognizers, so this could suggest that BUT English was trained on corpora with acoustic conditions that were more matched to the Tsonga corpus.
%

{\color{black}
Rows (14), (15), (16) show the results of fusing the supervised results in rows (8) to (11) with the unsupervised results in rows (1) to (7).
In all cases here the scores are simply averaged without weights.
Row (14) then shows the fused performance for supervised tokens (phonemes) when we fused the 4 supervised results in rows (8) to (11).
The fused result in row (14) is worse than row (8) on both corpora, which implies fusing the scores from unmatched supervised phoneme systems may actually hurt the performance.
In row (15), we combine row (14) with our unsupervised results in row (13) and manage to get some improvement, but on the English corpus it is still worse than the BUT English phonemes used alone in row (8).
In row (16) we fuse the results of the BUT English phoneme recognizer in row (8) alone with the unsupervised results from row (13) and get the best performance on the English corpus, and reasonably well performance on the Tsonga corpus.
By comparing row (16) to rows (13) and (8), we verify that the proposed unsupervised tokens and features can be complementary to supervised phonemes.
By comparing row (16) to row (14), we can further draw the conclusion that the improvement gain by combining a matched supervised phoneme system with our unsupervised system can actually be better than fusing it with other unmatched supervised systems.
}

%
In rows (17) to (21), we took the weighted average of the results in row (12) to (16) with weights tuned on a development query set. 
By comparing the results in rows (12) to (17), rows (13) to (18), rows (14) to (19), it is possible to achieve slightly better performance with weights tuned on a development.
By comparing rows (15), (16) to rows (20), (21), we see row (16) is better than row (15) for English because row (15) included the scores from three unmatched language phones which are harmful.
However, row (20) is approximately the same as row (21), meaning lower weights were assigned to the unmatched phones in row (20).
On the other hand, by comparing rows (16) to (21), the overall performance can be approximately the same even without a development set when we properly fuse unsupervised scores with the supervised scores.

%
%



\begin{table}[tb]
\caption {Spoken term detection performance in mean average precision.\label{tab:x}}
\begin{center}\begin{tabular}{|m{2cm}|l|l|r|r|}
\hline
\multirow{2}{*}{method}      & \multicolumn{2}{c|}{\multirow{2}{*}{index}} & \multicolumn{2}{c|}{MAP(\%)}                        \\ \cline{4-5} 
                             & \multicolumn{2}{c|}{}                       & \multicolumn{1}{c|}{Eng.} & \multicolumn{1}{c|}{Tso.} \\ \hline
\multirow{5}{*}{token-level DTW}  & (1) & TOK-1st, MR-0                  & 12.49 & 19.00 \\ \cline{2-5} 
                                  & (2) & TOK-1st, MR-1                  & 13.98 & 21.27 \\ \cline{2-5} 
                                  & (3) & TOK-1st, MR-2                  & 13.42 & 24.17 \\ \cline{2-5} 
                                  & (4) & TOK-2nd, MR-0                  & 10.37 & 25.58 \\ \cline{2-5} 
                                  & (5) & TOK-2nd, MR-1                  & 14.51 & 25.44 \\ \hline
\multirow{2}{*}{frame-level DTW}  & (6) & MFCC                           & 11.08 &  8.96 \\ \cline{2-5} 
                                  & (7) & BNF-1st, MR-1                  & 13.39 & 28.71 \\ \hline
\multirow{4}{*}{supervised phones}& (8) & BUT English                    & 16.76 & 19.85 \\ \cline{2-5} 
                                  & (9) & BUT Hungarian                  & 11.67 & 11.62 \\ \cline{2-5}
                                  & (10)& BUT Czech                      & 10.12 & 10.66 \\ \cline{2-5}
                                  & (11)& BUT Russian                    & 12.77 &  2.54 \\ \hline
fusion(I):                        & (12)& (1)-(5)                        & 15.28 & 26.17\\ \cline{2-5} 
token+frame                       & (13)& (1)-(7)                        & 18.01 & 26.33\\ \hline
fusion(II):                       & (14)& (8)-(11)                       & 14.28 &  8.30\\ \cline{2-5}
unsuper.+super.                   & (15)& (1)-(11)                       & 15.72 & 28.28\\ \cline{2-5}
                                  & (16)& (1)-(8)                        & 19.14 & 27.63\\ \hline
                                  & (17)& (1)-(5)  with dev.             & 17.89 & 26.00\\ \cline{2-5}
fusion(III):                      & (18)& (1)-(7)  with dev.             & 19.16 & 26.43\\ \cline{2-5}
with dev. set                     & (19)& (8)-(11) with dev.             & 15.01 & 17.18\\ \cline{2-5}
                                  & (20)& (1)-(11) with dev.             & 19.03 & 26.15\\ \cline{2-5}
                                  & (21)& (1)-(8)  with dev.             & 19.10 & 26.17\\ \hline
\end{tabular}
\end{center}
\end{table}

\subsection{Visualization of Discovered Tokens on TIMIT}\label{sec:vis}
To gain insight regarding what the discovered tokens really are, we applied the MAT on the TIMIT training set. 
We configured our MAT with the granularity setting of $m={3, 5, 7, 9}$, and $n={50, 100, 300, 500}$, with Mutual Reinforcement (MR) performed on these acoustic tokens for up to 3 iterations MR $={0, 1, 2, 3}$.
In other words we trained a total of 4$\times$4$\times$4 $= 64$ acoustic token sets.
For each token set we visualized the mapping from the tokens to English phonemes/words. 
By the time alignment between the realizations of the unsupervised tokens and those for the ground truth phonemes/words, we could show the co-occurrence of the discovered tokens and the ground truth phoneme/words by 2D scatter plots.
The SA sentences were removed from the TIMIT training set, so a huge proportion of words appear only once in the corpus.
In these figures, every point represented more than 100 realizations of a token on the vertical scale whose central frame belonged to the realization of an English phoneme on the horizontal scale.
%
Some selected results are presented below.

%
%
%
%

\begin{figure}[tb]
    \centering
    \includegraphics[width=0.8\columnwidth]{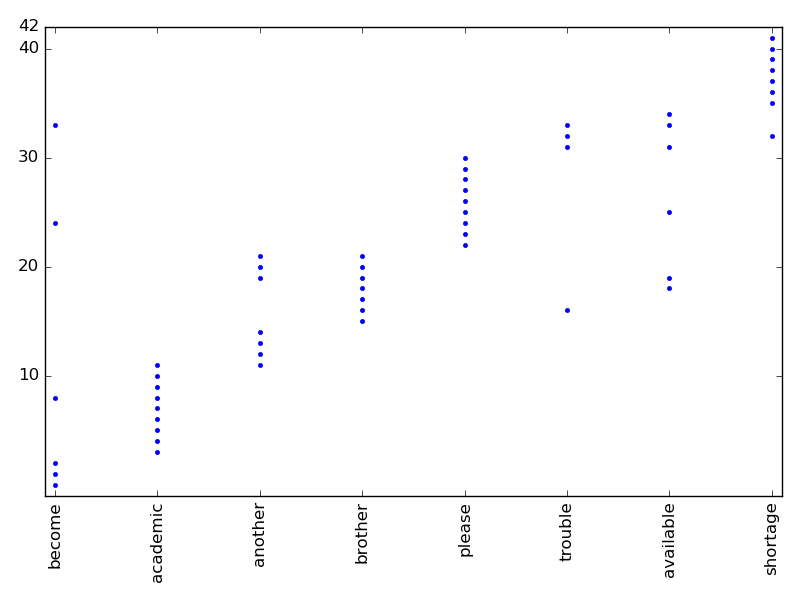}
    \caption{Mapping from the token set with $\psi=(5, 100)$ and MR $=1$ to TIMIT words}
    \label{fig:map_w}
\end{figure}

\begin{figure}[tb]
    \centering
    \includegraphics[width=0.8\columnwidth]{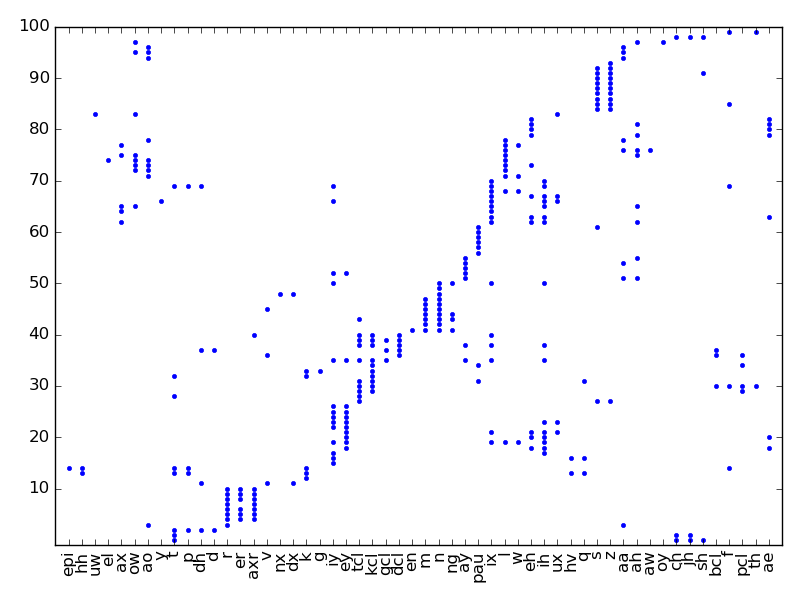}
    \caption{Mapping from the token set $\psi=(5, 100)$ and MR $=1$ to TIMIT phonemes}
    \label{fig:map_p}
\end{figure}

The results for the token set $\psi=(5, 100)$ and MR $=1$ with respect to English words are shown in Fig. \ref{fig:map_w}.
Here we selected the several words from the TIMIT training set and listed them on the x-axis. 
Each point in the figure represented 4 or more realizations of a discovered token on the y-axis whose central frame belonged to the word on the x-axis.
The tokens on the y-axis were ordered in such a way that in most cases the tokens appearing in an English word were grouped together.
Only those tokens appearing in these selected words are listed on the y-axis.
As a result, only 42 out of the 100 were present on the y-axis.
From Fig. \ref{fig:map_w}, we can observe that the mapping was quite clean, every word was represented by a small number of tokens and words with same suffixes share tokens, like (``another'' and ``brother''), (``trouble'' and ``available'').
%
%
%
%
We mapped the same token set used in Fig. \ref{fig:map_w} to all the English phonemes in TIMIT in Fig. \ref{fig:map_p}, but here all the 100 distinct tokens were listed.
The token order on the y-axis was again organized in such a way that distinct tokens appearing for the same phoneme were grouped together. 
As a results, the token order in Fig. \ref{fig:map_p} is different from that in Fig. \ref{fig:map_w}.
From Fig. \ref{fig:map_p}, we can find similar patterns for some phoneme pairs like (``s'' and ``z''), (``m'' and ``n'').
This implies that the token sets had successfully preserved the acoustic similarities between phonemes that sound similar.

\begin{figure}[tb]
    \centering
    \includegraphics[width=0.8\columnwidth]{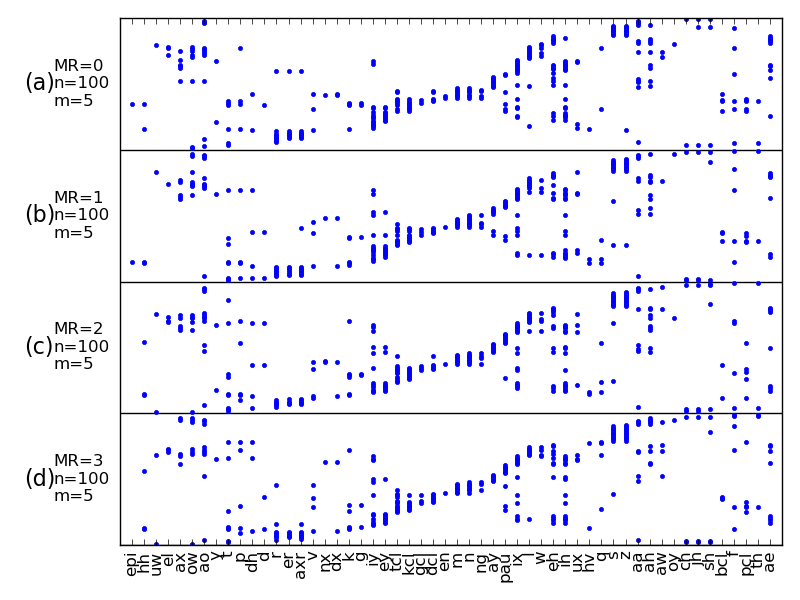}
    \caption{Mapping from tokens to TIMIT phonemes while sweeping the number of MR iterations from 0 to 3 and fixing $m=5$ and $n=100$.}
    \label{fig:map_r}
\end{figure}

\begin{figure}[tb]
    \centering
    \includegraphics[width=0.8\columnwidth]{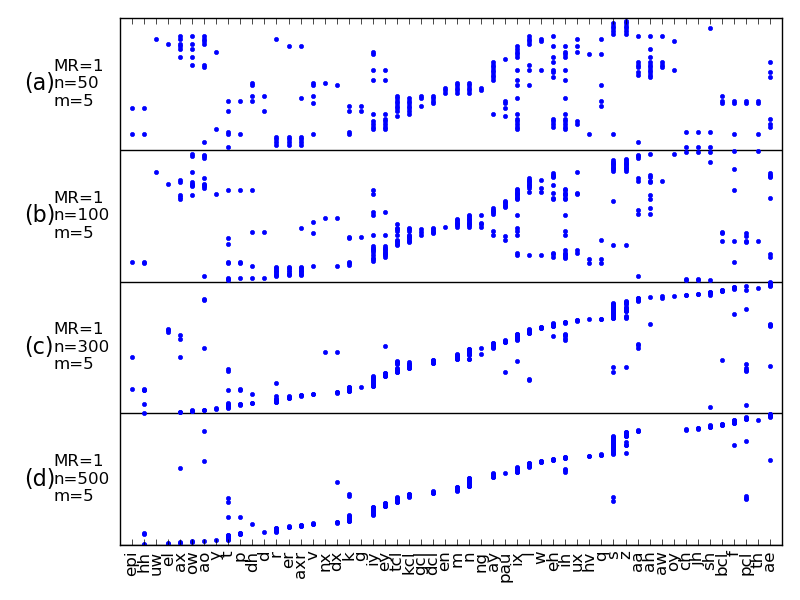}
    \caption{Mapping from tokens to TIMIT phonemes while sweeping the number of HMMs $n={50,100,300,500}$ and fixing $m=5$ and MR $=1$.}
    \label{fig:map_n}
\end{figure}

\begin{figure}[tb]
    \centering
    \includegraphics[width=0.8\columnwidth]{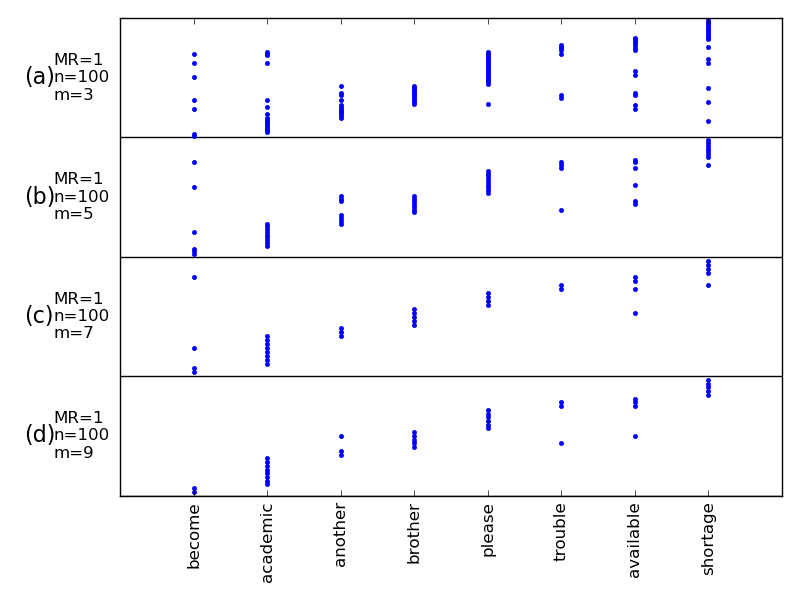}
    \caption{Mapping from tokens to English words while sweeping the length of token HMMs $m={3,5,7,9}$ and fixing $n=100$ and MR $=1$.}
    \label{fig:map_m}
\end{figure}

To further examine the effect of different values of $m$, $n$, and MR, we selected several tokens sets for comparison in Fig. \ref{fig:map_r}, Fig. \ref{fig:map_n} and Fig. \ref{fig:map_m}.
In each of these figures, we fixed the x-axis first so the subfigures for different parameters can be compared. 
The x-axis is sorted roughly based on acoustic similarity.
We then sorted the y axis of each subfigure roughly based on the acoustic similarity so that those tokens that sound similar were located closely.
In these figures, the y-axis of the subfigures were compressed to the same length so the general trend can be easily observed.

In Fig. \ref{fig:map_r}, we selected the token set with $\psi=(5, 100)$ and tracked its change over the MR iterations by comparing it to the English phonemes.
Every MR iteration is equivalent to retraining the tokens from scratch using a different initialization derived from the previous iteration.
In each MR iteration, a new set of token was generated, so the y-axis of each subfigure corresponded to a new set of tokens.
From Fig. \ref{fig:map_r}, we can see that the tokens converged to a relatively similar structure although through different number of MR iterations.
%
%
In all iterations, from Fig. \ref{fig:map_r} (a) to Fig. \ref{fig:map_r} (d) the subfigures changed only slightly.
This more or less supported our claim earlier that MR$=1$ would suffice under most situations.

In Fig. \ref{fig:map_n}, we focus the comparison on the different number of distinct tokens $n={50, 100, 300, 500}$ while fixing $m=5$ and MR $=1$ with respect to English phonemes.
In Fig. \ref{fig:map_n} (a) at $n=50$, the correspondence was more confused since there were not enough tokens to represent the 61 TIMIT phonemes, this means the acoustic tokens were underfitting in this situation.
As we increased $n$ to 500, the correspondence became cleaner as overfitting began to happen.
Fig. \ref{fig:map_n} depicts the difficulty in selecting a suitable $n$ on an unknown corpus.

\begin{figure}[tb]
    \centering
    \includegraphics[width=0.8\columnwidth]{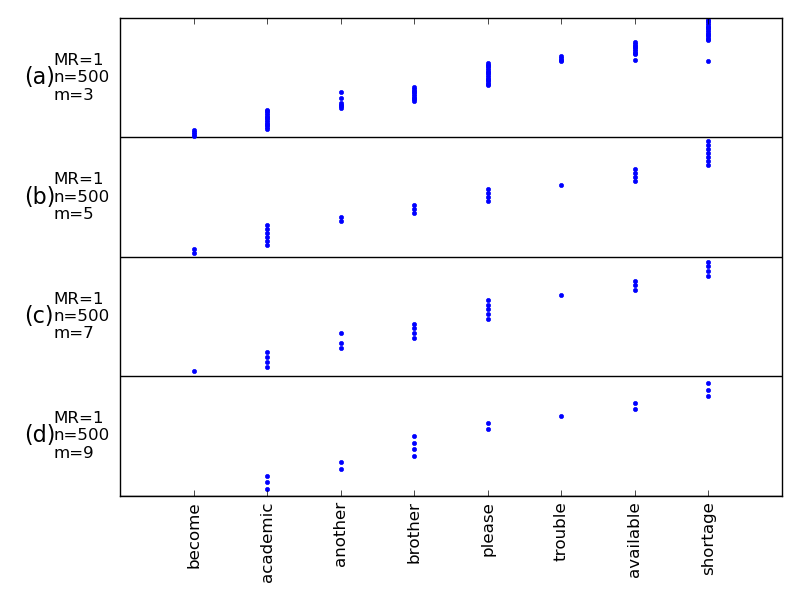}
    \caption{Mapping from tokens to English words while sweeping the length of token HMMs $m={3,5,7,9}$ and fixing $n=500$ and MR $=1$.}
    \label{fig:map_m_500}
\end{figure}

In Fig. \ref{fig:map_m}, we focus the comparison on the number of states in each token HMM, $m={3, 5, 7, 9}$, while fixing $n=100$ and MR $=1$ with respect to English words.
Note that Fig. \ref{fig:map_m} (b) is Fig. \ref{fig:map_w} with the vertical scale compressed.
We see with shorter tokens at $m=3$ in Fig. \ref{fig:map_m} (a), each English word was composed of more acoustic tokens.
However, as we increase the length of the acoustic tokens by increasing $m$ up to 9 in Fig. \ref{fig:map_m} (d), the English words became composed of fewer tokens.
{\color{black}
The same trend can be observed in Fig. \ref{fig:map_m_500}, where we sweep $m={3, 5, 7, 9}$, while fixing $n=500$ and MR $=1$. }

\begin{figure*}[th!]
    \centering
    \includegraphics[width=1.0\textwidth]{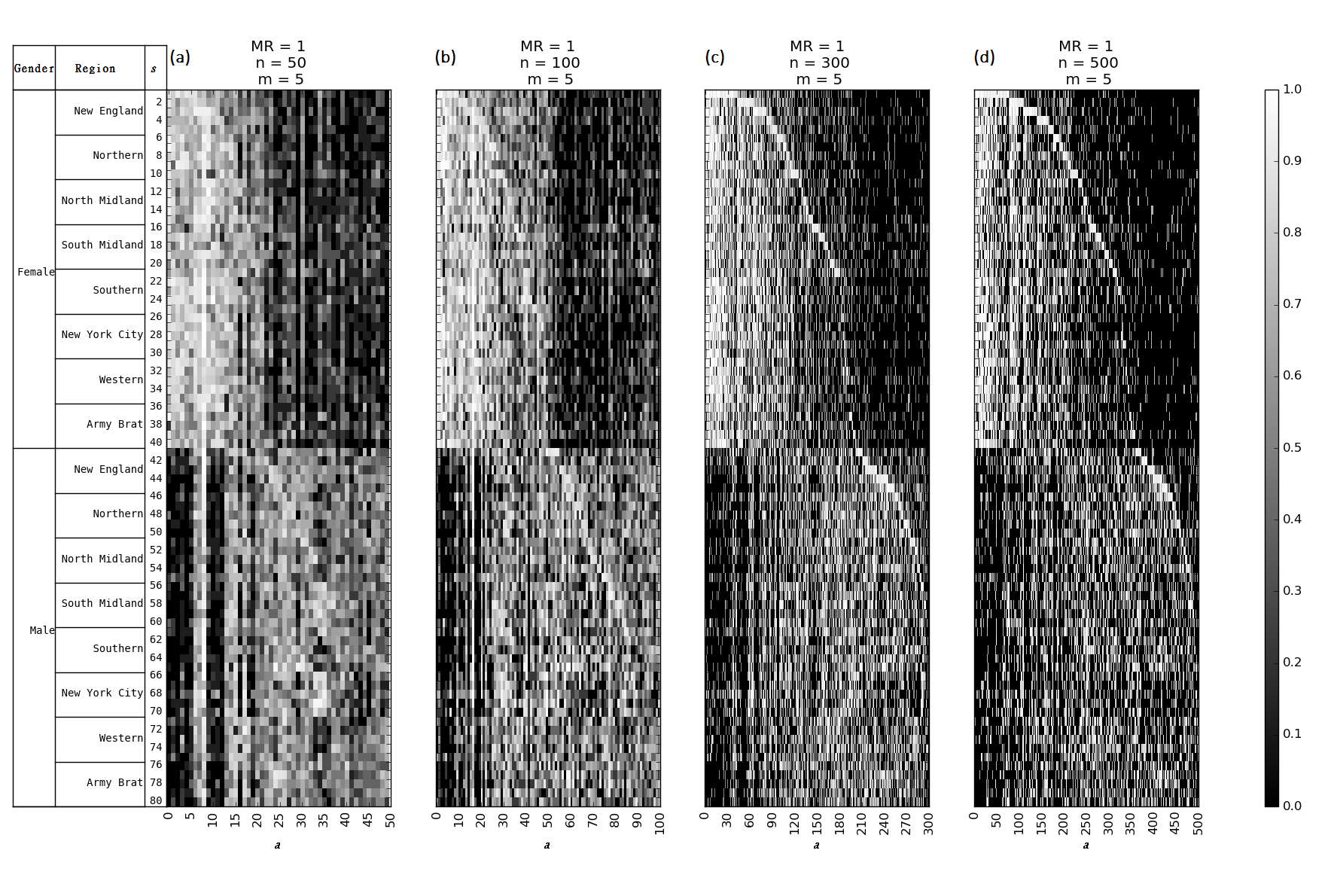}
   	\caption{The TIMIT speakers were sorted by their gender and dialect regions on the vertical scale ($s=1,...,80$), while the discovered acoustic tokens on the horizontal scale ($a=1,...,n$) are sorted based on the speakers for MR$=1$, $m=5$, $n=50, 100, 300, 500$ in (a)(b)(c)(d) to illustrate how different acoustic tokens were spoken by different speakers.}
    \label{fig:dialect}
\end{figure*}

{\color{black}
In Fig. \ref{fig:dialect}, we visualize how different acoustic tokens were spoken by different speakers at different phonetic granularity $n$.
We revisit the acoustic token sets in Fig. \ref{fig:map_n} by choosing $m=5$, MR $=1$ and $n={50, 100, 300, 500}$. 
This gives a total of 4 sets of acoustic tokens.
The speakers of the TIMIT dataset was composed of female and male speakers from 8 dialect regions in the United States of America: New England, Northern, North Midland, South Midland, Southern, New York City, Western, and Army Brat.
5 male and 5 female in the TIMIT dataset were randomly sampled from each of the 8 dialect regions, giving a total of 80, arranged on the vertical axis of Fig. \ref{fig:dialect}, as marked on the left, and indexed by $s=1,2,...,80$.
The the horizontal axes of the subfigures (a)(b)(c)(d) in Fig. \ref{fig:dialect} corresponds to the acoustic tokens indexed by $a=1,2,...,n$, $n = 50, 100, 300, 500$ respectively.
Let $c_{s,a}$ denote the count of acoustic token $a$ spoken by speaker $s$.
These counts were represented by a function $1 - \exp{(-\beta_n c_{s,a})}$ which gave a normalized occurrence between 0 and 1 shown as the intensity in the subfigures in Fig. \ref{fig:dialect}, where and $\beta_n$ was a normalization constant to equalize the average intensity across the 4 subfigures.
The occurrences were normalized for the purpose of presentation. 
%
%
If we did not normalize the occurrences, Fig. \ref{fig:dialect} (a) would be too bright and Fig. \ref{fig:dialect} (d) would be too dark, and cannot be shown on the same scale.
%

The acoustic tokens in the horizontal axis in each subfigure was first sorted based on the total counts in each subfigure, $\sum_s c_{s,a}$, then the order was adjusted in the following manner.
Let $l_s$ be the set of all the acoustic tokens $a$ spoken by speaker $s$ with $c_{s,a}$ exceeding a small threshold, referred to as frequently spoken tokens for speaker $s$. Let $L_s = \bigcup_{x=1}^s l_x$, which stands for all frequently spoken tokens counted from speaker 1 to $s$.
Starting with $s=1$ and $L_1=l_1$ we examined the extra frequently spoken tokens for the next speaker $L_{s+1}\setminus L_{s}$ (set difference, those tokens frequently spoken by speaker $s+1$, but not by speakers 1 to $s$). 
The tokens in $L_1=l_1$ are the first set on the horizontal axis $a=1$ up to $a=l_1$.
The next set are those in $L_2\setminus L_1$, followed by $L_3\setminus L_2$, and so on.
So we simply added new tokens frequently spoken by the next speaker one by one until $s = 80$.
This is the way the $n$ tokens were sorted on the horizontal axis in the 4 subfigures.
If $L_{s+1}\setminus L_{s}$ is an empty set, then speaker $s+1$ only used the tokens frequently spoken by the previous $s$ speakers, which means the speaker did not have extra speaker specific tokens.
If $L_{s+1}\setminus L_{s}$ is not an empty set, then speaker $s+1$ used extra frequently spoken tokens not frequently spoken by the previous speakers, which implies there were speaker specific tokens.
For example, in Fig. \ref{fig:dialect} (d) for $n=500$, the first white horizontal bar at $s=1$ are those tokens in $L_1=l_1$, and the next white horizontal bar at $s=2$ are those in $L_2\setminus L_1$, and so on.
When new frequently spoken acoustic tokens were added with every newly considered speaker, a concentrated white curve was formed on Fig. \ref{fig:dialect} (c) and (d).
The slope of these curves depend on the the number of newly added frequently spoken tokens per speaker.
The steeper the curves, the fewer extra frequently spoken tokens were added per speaker, which implies that the newly considered speaker was more similar to the existing speakers.

From Fig. \ref{fig:dialect}, we can see that the female speaker (upper half) and male speakers (lower half) have very different acoustic token distributions.
At smaller $n$ in Fig. \ref{fig:dialect} (a)(b) we see more tokens were shared by both genders, while at larger $n$ in Fig. \ref{fig:dialect} (c)(d), the distribution is more separated.
This implies that at larger $n$, the MAT discovered enough number of tokens to make them gender dependent.
A similar trend can be observed for speaker specific information.
At smaller $n$ in Fig. \ref{fig:dialect} (a) , the first speaker for each gender ($s=1$ and $s=41$) used up almost all the acoustic tokens for each gender, so the acoustic token distributions look very similar for all speakers in each gender.
At larger $n$ in Fig. \ref{fig:dialect} (c)(d) , we have more speaker specific acoustic tokens, so many newly considered speakers added new frequently spoken tokens to the existing frequently spoken acoustic token list and this created the concentrated white curves that we see in the figure as mentioned above.
This implies that at higher $n$, the MAT discovered speaker specific acoustic tokens.
It is also worth noting that by observing the slope for the concentrated white curves, we can barely visualize the impact of dialect regions.
In Fig. \ref{fig:dialect} (d) for $n=500$, the slope looks relatively lower at a few transitions between dialect regions, and relatively higher within a few dialect regions. 
This implies the speakers within each dialect region are more similar to each other.
}


\section{Conclusion}\label{sec:VI}
In this paper we propose an iterative deep learning framework, the Multi-granular Acoustic Tokenizing Deep Neural Network (MAT-DNN), to iteratively learn high quality frame-level features and multi-granular acoustic token sets from a given audio corpus in a completely unsupervised way. 
These features and tokens were evaluated by the metrics and corpora defined in the Zero Resource Speech Challenge in Interspeech 2015. 
We also tested the token sets and frame-level features with a query-by-example spoken term detection task.
When the information obtained from different token sets and frame-level features were fused in the spoken term detection experiments, good initial results were obtained. 
The unsupervised tokens were competitive when compared to supervised phoneme recognizer from four other languages on the task of STD.
The best results were obtained when we fuse the supervised and the proposed unsupervised systems, which implies the MAT-DNN is complementary to supervised systems.
In addition, we performed a series of visualization experiments on TIMIT to  understand what tokens with different granularities represent.
We hope that these results serve as good references for future investigations.


\bibliographystyle{IEEEbib}
\bibliography{mycap}

\begin{thebibliography}{10}

\bibitem{hinton2012deep}
Geoffrey Hinton, Li~Deng, Dong Yu, George~E Dahl, Abdel-rahman Mohamed, Navdeep
  Jaitly, Andrew Senior, Vincent Vanhoucke, Patrick Nguyen, Tara~N Sainath,
  et~al.,
\newblock ``{D}eep {N}eural {N}etworks {F}or {A}coustic {M}odeling {I}n
  {S}peech {R}ecognition: {T}he {S}hared {V}iews {O}f {F}our {R}esearch
  {G}roups,''
\newblock {\em Signal Processing Magazine, IEEE}, vol. 29, no. 6, pp. 82--97,
  2012.

\bibitem{szoke2015copingwith}
Igor Sz{\"o}ke, Miroslav Sk{\'a}cel, Luk{\'a}{\v{s}} Burget, and Jan
  {\v{C}}ernock{\`y},
\newblock ``Coping with channel mismatch in query-by-example-but quesst 2014,''
\newblock in {\em Acoustics, Speech and Signal Processing (ICASSP), 2015 IEEE
  International Conference on}. IEEE, 2015, pp. 5838--5842.

\bibitem{leung2016toward}
Cheung-Chi Leung, Lei Wang, Haihua Xu, Jingyong Hou, Van~Tung Pham, Hang Lv,
  Lei Xie, Xiong Xiao, Chongjia Ni, Bin Ma, et~al.,
\newblock ``Toward high-performance language-independent query-by-example
  spoken term detection for mediaeval 2015: Post-evaluation analysis,''
\newblock in {\em Proc. INTERSPEECH}, 2016.

\bibitem{chen2016unsupervised}
Hongjie Chen, Cheung-Chi Leung, Lei Xie, Bin Ma, and Haizhou Li,
\newblock ``Unsupervised bottleneck features for low-resource query-by-example
  spoken term detection,''
\newblock in {\em Proc. INTERSPEECH}, 2016.

\bibitem{yang2014intrinsic}
Peng Yang, Cheung-Chi Leung, Lei Xie, Bin Ma, and Haizhou Li,
\newblock ``Intrinsic spectral analysis based on temporal context features for
  query-by-example spoken term detection.,''
\newblock in {\em INTERSPEECH}, 2014, pp. 1722--1726.

\bibitem{wang2015acoustic}
Haipeng Wang, Tan Lee, Cheung-Chi Leung, Bin Ma, and Haizhou Li,
\newblock ``Acoustic segment modeling with spectral clustering methods,''
\newblock {\em IEEE/ACM Transactions on Audio, Speech and Language Processing
  (TASLP)}, vol. 23, no. 2, pp. 264--277, 2015.

\bibitem{renshaw2015comparison}
Daniel Renshaw, Herman Kamper, Aren Jansen, and Sharon Goldwater,
\newblock ``A {C}omparison of {N}eural {N}etwork {M}ethods for {U}nsupervised
  {R}epresentation {L}earning on the {Z}ero {R}esource {S}peech {C}hallenge,''
\newblock in {\em Proceedings of Interspeech}, 2015.

\bibitem{zhang2013unsupervised}
Yaodong Zhang,
\newblock {\em Unsupervised {S}peech {P}rocessing with {A}pplications to
  query-by-example {S}poken {T}erm {D}etection},
\newblock Ph.D. thesis, Massachusetts Institute of Technology, 2013.

\bibitem{huang2015rapid}
Zhen Huang, Jinyu Li, Sabato~Marco Siniscalchi, I-Fan Chen, Ji~Wu, and Chin-Hui
  Lee,
\newblock ``Rapid adaptation for deep neural networks through multi-task
  learning.,''
\newblock in {\em Interspeech}, 2015, pp. 3625--3629.

\bibitem{jansen2012indexing}
Aren Jansen and Benjamin Van~Durme,
\newblock ``Indexing {R}aw {A}coustic {F}eatures {F}or {S}calable {Z}ero
  {R}esource {S}earch.,''
\newblock in {\em INTERSPEECH}, 2012.

\bibitem{jansen2011towards}
Aren Jansen and Kenneth Church,
\newblock ``Towards {U}nsupervised {T}raining {O}f {S}peaker {I}ndependent
  {A}coustic {M}odels.,''
\newblock in {\em INTERSPEECH}, 2011, pp. 1693--1692.

\bibitem{gish2009unsupervised}
Herbert Gish, Man-hung Siu, Arthur Chan, and William Belfield,
\newblock ``Unsupervised {T}raining {O}f {A}n {H}mm-based {S}peech {R}ecognizer
  {F}or {T}opic {C}lassification.,''
\newblock in {\em INTERSPEECH}, 2009, pp. 1935--1938.

\bibitem{chung2013unsupervised}
Cheng-Tao Chung, Chun-an Chan, and Lin-shan Lee,
\newblock ``Unsupervised {D}iscovery {O}f {L}inguistic {S}tructure {I}ncluding
  {T}wo-level {A}coustic {P}atterns {U}sing {T}hree {C}ascaded {S}tages {O}f
  {I}terative {O}ptimization,''
\newblock in {\em Acoustics, {S}peech {A}nd {S}ignal {P}rocessing (ICASSP),
  {2}013 {I}EEE {I}nternational {C}onference {O}n}. IEEE, 2013, pp. 8081--8085.

\bibitem{chung2014unsupervised}
Cheng-Tao Chung, Chun-an Chan, and Lin-shan Lee,
\newblock ``Unsupervised {S}poken {T}erm {D}etection {W}ith {S}poken {Q}ueries
  {B}y {M}ulti-level {A}coustic {P}atterns {W}ith {V}arying {M}odel
  {G}ranularity,''
\newblock in {\em Acoustics, {S}peech {A}nd {S}ignal {P}rocessing (ICASSP),
  {2}014 {I}EEE {I}nternational {C}onference {O}n}. IEEE, 2014.

\bibitem{chung2015enhancing}
Cheng-Tao Chung, Wei-Ning Hsu, Cheng-Yi Lee, and Lin-Shan Lee,
\newblock ``{E}nhancing {A}utomatically {D}iscovered {M}ulti-level {A}coustic
  {P}atterns{C}onsidering {C}ontext {C}onsistency with {A}pplications in
  {S}poken {T}erm {D}etection,''
\newblock in {\em Acoustics, {S}peech {A}nd {S}ignal {P}rocessing (ICASSP),
  {2}015 {I}EEE {I}nternational {C}onference {O}n}. IEEE, 2015.

\bibitem{li2013towards}
Yun-Chiao Li, Hung-yi Lee, Cheng-Tao Chung, Chun-an Chan, and Lin-shan Lee,
\newblock ``Towards unsupervised semantic retrieval of spoken content with
  query expansion based on automatically discovered acoustic patterns,''
\newblock in {\em Automatic Speech Recognition and Understanding (ASRU), 2013
  IEEE Workshop on}. IEEE, 2013, pp. 198--203.

\bibitem{chung2015iterative}
Cheng-Tao Chung, Cheng-Yu Tsai, Hsiang-Hung Lu, Chia-Hsiang Liu, Hung-yi Lee,
  and Lin-shan Lee,
\newblock ``{A}n {I}terative {D}eep {L}earning {F}ramework for {U}nsupervised
  {D}iscovery of {S}peech {F}eatures and {L}inguistic {U}nits with
  {A}pplications on {S}poken {T}erm {D}etection,''
\newblock in {\em Automatic {S}peech {R}ecognition {A}nd {U}nderstanding
  (ASRU), {2}015 {I}EEE {W}orkshop {O}n}. IEEE, 2015.

\bibitem{vu2014investigating}
Ngoc~Thang Vu, Jochen Weiner, and Tanja Schultz,
\newblock ``{I}nvestigating {T}he {L}earning {E}ffect {O}f {M}ultilingual
  {B}ottle-{N}eck {F}eatures {F}or {ASR},''
\newblock in {\em Fifteenth {A}nnual {C}onference {O}f {T}he {I}nternational
  {S}peech {C}ommunication {A}ssociation}, 2014.

\bibitem{vesely2012language}
Karel Vesely, Martin Karafi{\'a}t, Frantisek Grezl, Marcel Janda, and Ekaterina
  Egorova,
\newblock ``{T}he {L}anguage-{I}ndependent {B}ottleneck {F}eatures,''
\newblock in {\em Spoken {L}anguage {T}echnology {W}orkshop (SLT), {2}012
  {I}EEE}. IEEE, 2012, pp. 336--341.

\bibitem{versteegh2015zero}
Maarten Versteegh, Roland Thiolliere, Thomas Schatz, Xuan~Nga Cao, Xavier
  Anguera, Aren Jansen, and Emmanuel Dupoux,
\newblock ``The {Z}ero {R}esource {S}peech {C}hallenge 2015,''
\newblock in {\em Proc. of INTERSPEECH}, 2015.

\bibitem{lee2012nonparametric}
Chia-ying Lee and James Glass,
\newblock ``A {N}onparametric {B}ayesian {A}pproach {T}o {A}coustic {M}odel
  {D}iscovery,''
\newblock in {\em Proceedings {O}f {T}he {5}0th {A}nnual {M}eeting {O}f {T}he
  {A}ssociation {F}or {C}omputational {L}inguistics: {L}ong {P}apers-Volume
  {1}}. Association for Computational Linguistics, 2012, pp. 40--49.

\bibitem{siu2014unsupervised}
Man-hung Siu, Herbert Gish, Arthur Chan, William Belfield, and Steve Lowe,
\newblock ``{U}nsupervised {T}raining {O}f {A}n {HMM}-based {S}elf-{O}rganizing
  {U}nit {R}ecognizer {W}ith {A}pplications {T}o {T}opic {C}lassification {A}nd
  {K}eyword {D}iscovery,''
\newblock {\em Computer Speech \& Language}, vol. 28, no. 1, pp. 210--223,
  2014.

\bibitem{kamperunsupervised}
Herman Kamper, Micha Elsner, Aren Jansen, and Sharon Goldwater,
\newblock ``{U}nsupervised {N}eural {N}etwork {B}ased {F}eature {E}xtraction
  {U}sing {W}eak {T}op-{D}own {C}onstraints,''
\newblock .

\bibitem{levin2015segmental}
Keith Levin, Aren Jansen, and Benjamin Van~Durme,
\newblock ``{S}egmental {A}coustic {I}ndexing {F}or {Z}ero {R}esource {K}eyword
  {S}earch,''
\newblock in {\em Proc. {I}CASSP}, 2015.

\bibitem{lee1988segment}
Chin-Hui Lee, Frank~K Soong, and Bing-Hwang Juang,
\newblock ``A {S}egment {M}odel based {A}pproach to {S}peech {R}ecognition,''
\newblock in {\em Acoustics, Speech, and Signal Processing, 1988. ICASSP-88.,
  1988 International Conference on}. IEEE, 1988, pp. 501--541.

\bibitem{wang2011unsupervised}
Haipeng Wang, Tan Lee, and Cheung-Chi Leung,
\newblock ``Unsupervised spoken term detection with acoustic segment model,''
\newblock in {\em Speech Database and Assessments (Oriental COCOSDA), 2011
  International Conference on}. IEEE, 2011, pp. 106--111.

\bibitem{siu2010improved}
Man-Hung Siu, Herbert Gish, Arthur Chan, and William Belfield,
\newblock ``Improved {T}opic {C}lassification {A}nd {K}eyword {D}iscovery
  {U}sing {A}n {H}mm-based {S}peech {R}ecognizer {T}rained {W}ithout
  {S}upervision.,''
\newblock in {\em INTERSPEECH}, 2010, pp. 2838--2841.

\bibitem{creutz2007unsupervised}
Mathias Creutz and Krista Lagus,
\newblock ``Unsupervised {M}odels {F}or {M}orpheme {S}egmentation {A}nd
  {M}orphology {L}earning,''
\newblock {\em ACM Transactions on Speech and Language Processing (TSLP)}, vol.
  4, no. 1, pp. 3, 2007.

\bibitem{jansen2010towards}
Aren Jansen, Kenneth Church, and Hynek Hermansky,
\newblock ``Towards {S}poken {T}erm {D}iscovery {A}t {S}cale {W}ith {Z}ero
  {R}esources.,''
\newblock in {\em INTERSPEECH}, 2010, pp. 1676--1679.

\bibitem{couprie1997topological}
Michel Couprie and Gilles Bertrand,
\newblock ``{T}opological gray-scale {W}atershed {T}ransformation,''
\newblock in {\em Optical Science, Engineering and Instrumentation'97}.
  International Society for Optics and Photonics, 1997, pp. 136--146.

\bibitem{blei2003latent}
David~M Blei, Andrew~Y Ng, and Michael~I Jordan,
\newblock ``{L}atent {D}irichlet {A}llocation,''
\newblock {\em the Journal of machine Learning research}, vol. 3, pp.
  993--1022, 2003.

\bibitem{salakhutdinov2009deep}
Ruslan Salakhutdinov and Geoffrey~E Hinton,
\newblock ``{D}eep {B}oltzmann {M}achines,''
\newblock in {\em International {C}onference {O}n {A}rtificial {I}ntelligence
  {A}nd {S}tatistics}, 2009, pp. 448--455.

\bibitem{hochreiter1997long}
Sepp Hochreiter and J{\"u}rgen Schmidhuber,
\newblock ``{L}ong {S}hort-{T}erm {M}emory,''
\newblock {\em Neural computation}, vol. 9, no. 8, pp. 1735--1780, 1997.

\bibitem{kanagasundaram2011vector}
Ahilan Kanagasundaram, Robbie Vogt, David~B Dean, Sridha Sridharan, and
  Michael~W Mason,
\newblock ``{I}-vector {B}ased {S}peaker {R}ecognition {O}n {S}hort
  {U}tterances,''
\newblock in {\em Proceedings {O}f {T}he {1}2th {A}nnual {C}onference {O}f
  {T}he {I}nternational {S}peech {C}ommunication {A}ssociation}. International
  Speech Communication Association (ISCA), 2011, pp. 2341--2344.

\bibitem{hershey2007approximating}
John~R Hershey and Peder~A Olsen,
\newblock ``Approximating {T}he {K}ullback {L}eibler {D}ivergence {B}etween
  {G}aussian {M}ixture {M}odels,''
\newblock in {\em Acoustics, {S}peech {A}nd {S}ignal {P}rocessing, {2}007.
  {I}CASSP {2}007. {I}EEE {I}nternational {C}onference {O}n}. IEEE, 2007,
  vol.~4, pp. IV--317.

\bibitem{zhang2012resource}
Yaodong Zhang, Ruslan Salakhutdinov, Hung-An Chang, and James Glass,
\newblock ``Resource {C}onfigurable {S}poken {Q}uery {D}etection using {D}eep
  {B}oltzmann {M}achines,''
\newblock in {\em Acoustics, Speech and Signal Processing (ICASSP), 2012 IEEE
  International Conference on}. IEEE, 2012, pp. 5161--5164.

\bibitem{pitt2007buckeye}
Mark~A Pitt, Laura Dilley, Keith Johnson, Scott Kiesling, William Raymond,
  Elizabeth Hume, and Eric Fosler-Lussier,
\newblock ``Buckeye {C}orpus {O}f {C}onversational {S}peech (2nd {R}elease),''
\newblock {\em Columbus, OH: Department of Psychology, Ohio State University},
  2007.

\bibitem{chung2014zero}
Cheng-Tao Chung,
\newblock ``zrst,'' {https://github.com/C2Tao/zrst}, 2014.

\bibitem{young1997htk}
Steve Young, Gunnar Evermann, Mark Gales, Thomas Hain, Dan Kershaw, Xunying
  Liu, Gareth Moore, Julian Odell, Dave Ollason, Dan Povey, et~al.,
\newblock {\em The {HTK} {B}ook}, vol.~2,
\newblock Entropic Cambridge Research Laboratory Cambridge, 1997.

\bibitem{stolcke2002srilm}
Andreas Stolcke et~al.,
\newblock ``{SRILM}-an {E}xtensible {L}anguage {M}odeling {T}oolkit.,''
\newblock in {\em INTERSPEECH}, 2002.

\bibitem{Povey_ASRU2011}
Daniel Povey, Arnab Ghoshal, Gilles Boulianne, Lukas Burget, Ondrej Glembek,
  Nagendra Goel, Mirko Hannemann, Petr Motlicek, Yanmin Qian, Petr Schwarz, Jan
  Silovsky, Georg Stemmer, and Karel Vesely,
\newblock ``The {K}aldi {S}peech {R}ecognition {T}oolkit,''
\newblock in {\em IEEE 2011 Workshop on Automatic Speech Recognition and
  Understanding}. Dec. 2011, IEEE Signal Processing Society,
\newblock IEEE Catalog No.: CFP11SRW-USB.

\bibitem{jia2014caffe}
Yangqing Jia, Evan Shelhamer, Jeff Donahue, Sergey Karayev, Jonathan Long, Ross
  Girshick, Sergio Guadarrama, and Trevor Darrell,
\newblock ``Caffe: {C}onvolutional {A}rchitecture for {F}ast {F}eature
  {E}mbedding,''
\newblock {\em arXiv preprint arXiv:1408.5093}, 2014.

\bibitem{schatz2013evaluating}
Thomas Schatz, Vijayaditya Peddinti, Francis Bach, Aren Jansen, Hynek
  Hermansky, and Emmanuel Dupoux,
\newblock ``Evaluating {S}peech {F}eatures with the {M}inimal-{P}air {ABX}
  task: {A}nalysis of the {C}lassical {MFC/PLP} pipeline,''
\newblock in {\em INTERSPEECH 2013: 14th Annual Conference of the International
  Speech Communication Association}, 2013, pp. 1--5.

\bibitem{schatz2014evaluating}
Thomas Schatz, Vijayaditya Peddinti, Xuan-Nga Cao, Francis~R Bach, Hynek
  Hermansky, and Emmanuel Dupoux,
\newblock ``Evaluating {S}peech {F}eatures with the {M}inimal-{P}air {ABX} task
  (ii): {R}esistance to {N}oise.,''
\newblock 2014.

\bibitem{philbin2007object}
James Philbin, Ondrej Chum, Michael Isard, Josef Sivic, and Andrew Zisserman,
\newblock ``Object {R}etrieval {W}ith {L}arge {V}ocabularies {A}nd {F}ast
  {S}patial {M}atching,''
\newblock in {\em Computer {V}ision {A}nd {P}attern {R}ecognition, {2}007.
  {C}VPR'07. {I}EEE {C}onference {O}n}. IEEE, 2007, pp. 1--8.

\bibitem{yue2007support}
Yisong Yue, Thomas Finley, Filip Radlinski, and Thorsten Joachims,
\newblock ``A {S}upport {V}ector {M}ethod {F}or {O}ptimizing {A}verage
  {P}recision,''
\newblock in {\em Proceedings {O}f {T}he {3}0th {A}nnual {I}nternational {A}CM
  {S}IGIR {C}onference {O}n {R}esearch {A}nd {D}evelopment {I}n {I}nformation
  {R}etrieval}. ACM, 2007, pp. 271--278.

\bibitem{muller2007dynamic}
Meinard M{\"u}ller,
\newblock ``Dynamic {T}ime {W}arping,''
\newblock {\em Information retrieval for music and motion}, pp. 69--84, 2007.

\bibitem{thiolliere2015hybrid}
Roland Thiolliere, Ewan Dunbar, Gabriel Synnaeve, Maarten Versteegh, and
  Emmanuel Dupoux,
\newblock ``A {H}ybrid {D}ynamic {T}ime {W}arping-deep {N}eural {N}etwork
  {A}rchitecture for {U}nsupervised {A}coustic {M}odeling,''
\newblock in {\em Sixteenth Annual Conference of the International Speech
  Communication Association}. Citeseer, 2015.

\bibitem{badino2015discovering}
Leonardo Badino, Alessio Mereta, and Lorenzo Rosasco,
\newblock ``Discovering {D}iscrete {S}ubword units with {B}inarized
  {A}utoencoders and {H}idden-markov-model {E}ncoders,''
\newblock in {\em Proceedings of Interspeech}, 2015.

\bibitem{chen2015parallel}
Hongjie Chen, Cheung-Chi Leung, Lei Xie, Bin Ma, and Haizhou Li,
\newblock ``{P}arallel {I}nference of {D}irichlet {P}rocess {G}aussian
  {M}ixture {M}odels for {U}nsupervised {A}coustic {M}odeling: A feasibility
  study,''
\newblock in {\em Proceedings of Interspeech}, 2015.

\bibitem{baljekar2015using}
Pallavi Baljekar, Sunayana Sitaram, Prasanna~Kumar Muthukumar, and A~Black,
\newblock ``{U}sing {A}rticulatory {F}eatures and {I}nferred {P}honological
  {S}egments in {Z}ero {R}esource {S}peech {P}rocessing,''
\newblock in {\em Proceedings of Interspeech}, 2015.

\bibitem{rasanen2015unsupervised}
Okko R{\"a}s{\"a}nen, Gabriel Doyle, and Michael~C Frank,
\newblock ``{U}nsupervised {W}ord {D}iscovery from {S}peech using {A}utomatic
  {S}egmentation into {S}yllable-like {U}nits,''
\newblock in {\em Proceedings of Interspeech}, 2015.

\bibitem{lyzinski2015evaluation}
Vince Lyzinski, Gregory Sell, and Aren Jansen,
\newblock ``An {E}valuation of {G}raph {C}lustering {M}ethods for
  {U}nsupervised {T}erm {D}iscovery,''
\newblock in {\em Proceedings of Interspeech}, 2015.

\bibitem{ludusan2014bridging}
Bogdan Ludusan, Maarten Versteegh, Aren Jansen, Guillaume Gravier, Xuan-Nga
  Cao, Mark Johnson, and Emmanuel Dupoux,
\newblock ``{B}ridging {T}he {G}ap {B}etween {S}peech {T}echnology {A}nd
  {N}atural {L}anguage {P}rocessing: {A}n {E}valuation {T}oolbox {F}or {T}erm
  {D}iscovery {S}ystems,''
\newblock in {\em Language {R}esources {A}nd {E}valuation {C}onference}, 2014.

\bibitem{jansen2011efficient}
Aren Jansen and Benjamin Van~Durme,
\newblock ``{E}fficient {S}poken {T}erm {D}iscovery {U}sing {R}andomized
  {A}lgorithms,''
\newblock in {\em Automatic {S}peech {R}ecognition {A}nd {U}nderstanding
  (ASRU), {2}011 {I}EEE {W}orkshop {O}n}. IEEE, 2011, pp. 401--406.

\bibitem{johnson2006adaptor}
Mark Johnson, Thomas~L Griffiths, and Sharon Goldwater,
\newblock ``{A}daptor {G}rammars: A framework for {S}pecifying {C}ompositional
  {N}onparametric {B}ayesian {M}odels,''
\newblock in {\em Advances in neural information processing systems}, 2006, pp.
  641--648.

\bibitem{matejka2005phonotactic}
Pavel Matejka and Petr Schwarz,
\newblock ``Phonotactic {L}anguage {I}dentification using {H}igh {Q}uality
  {P}honeme {R}ecognition.,''
\newblock .

\end{thebibliography}

\end{document}